%% file: main.tex
\documentclass[acmlarge, review, imwut, anonymous=false]{acmart}

\usepackage[caption=false]{subfig}
\usepackage[export]{adjustbox}
\usepackage{makecell}
\usepackage{amsmath,amsfonts}
\usepackage{algorithmic}
\usepackage{graphicx}
\usepackage{textcomp}
\usepackage{xcolor}
\usepackage{multirow}
\usepackage{booktabs}
\usepackage{hhline}
\usepackage[shortlabels]{enumitem}
\usepackage{soul}
\usepackage{fancyhdr}   
\usepackage{calc}
\usepackage{atbegshi}
\usepackage{tikz}
\usepackage{bm}

\usepackage{geometry}
\usepackage[utf8]{inputenc}
\usepackage[ruled]{algorithm2e}
\usepackage{appendix}

\usepackage{color, colortbl}  

\AtBeginDocument{%
  }

\copyrightyear{2023}
\acmYear{2023}
\setcopyright{rightsretained}

\definecolor{rubinered}{HTML}{CE0058}

\definecolor{green}{rgb}{0.0, 0.65, 0.31}
\definecolor{yellow}{rgb}{0.7, 0.4, 0}

\definecolor{bleudefrance}{rgb}{0.19, 0.55, 0.91}

\definecolor{ao(english)}{rgb}{0.0, 0.5, 0.0}

\definecolor{violet}{HTML}{6a51a3}

\definecolor{awesome}{rgb}{1.0, 0.13, 0.32}

\begin{document}

\title[IMUGPT 2.0]{IMUGPT 2.0: Language-Based Cross Modality Transfer for Sensor-Based Human Activity Recognition}

\author{Zikang Leng}
\email{zleng7@gatech.edu}
\orcid{0000-0001-6789-4780}
\affiliation{%
  \institution{College of Computing, Georgia Institute of Technology}
  \city{Atlanta, Georgia}
  \country{USA}}

\author{Amitrajit Bhattacharjee}
\email{amit.bh@gatech.edu}
\orcid{0009-0000-4713-1786}
\affiliation{%
  \institution{School of Interactive Computing, Georgia Institute of Technology}
  \city{Atlanta, Georgia}
  \country{USA}}

\author{Hrudhai Rajasekhar}
\email{hrajasekhar3@gatech.edu}
\orcid{0009-0008-4790-3588}
\affiliation{%
  \institution{School of Interactive Computing, Georgia Institute of Technology}
  \city{Atlanta, Georgia}
  \country{USA}}

\author{Lizhe Zhang}
\email{lzhang762@gatech.edu}
\orcid{0009-0008-4609-9436}
\affiliation{%
  \institution{College of Engineering , Georgia Institute of Technology}
  \city{Atlanta, Georgia}
  \country{USA}}

\author{Elizabeth Bruda}
\email{ebruda3@gatech.edu}
\orcid{0009-0003-1032-5933}
\affiliation{%
  \institution{College of Computing, Georgia Institute of Technology}
  \city{Atlanta, Georgia}
  \country{USA}}
  
\author{Hyeokhyen Kwon}
\email{hyeokhyen.kwon@emory.edu}
\orcid{0000-0002-5693-3278}
\affiliation{%
  \institution{Department of Biomedical Informatics, Emory University}
  \city{Atlanta, Georgia}
  \country{USA}}
  
\author{Thomas Plötz}
\email{thomas.ploetz@gatech.edu}
\orcid{0000-0002-1243-7563}
\affiliation{%
  \institution{School of Interactive Computing, Georgia Institute of Technology}
  \city{Atlanta, Georgia}
  \country{USA}}

\renewcommand{\shortauthors}{Leng et al.}

\begin{abstract}
\input{abstract/main}
\end{abstract}

\ccsdesc[500]{Human-centered computing~Ubiquitous and mobile computing}
\ccsdesc[500]{Computing methodologies~Artificial intelligence}

\keywords{Wearables; Activity recognition; Virtual IMU Data; Motion Synthesis; LLM}


\input{figure/general_flow}

\maketitle

\pagestyle{fancy}
\fancyhf{}
\renewcommand{\headrulewidth}{0pt}
\AtBeginShipout{\AtBeginShipoutAddToBox{%
  \begin{tikzpicture}[remember picture, overlay, red]
    \node[anchor=south, font=\LARGE] at ([yshift=15mm]current page.south) {This manuscript is under review. Please write to zleng7@gatech.edu for up-to-date information};
  \end{tikzpicture}%
}}

\input{sections/intro/main}

\input{sections/related_work/main}

\input{sections/method/main}

\input{sections/experiment/main}


\input{sections/discussion/main}

\input{sections/conclusion/main}

\bibliographystyle{ACM-Reference-Format}
\bibliography{./bibs/virtual_imu,./bibs/motion_synthesis, ./bibs/har, ./bibs/nlp}

\newpage
\input{appendix/main}


\end{document}

%% file: abstract/main.tex
\noindent
One of the primary challenges in the field of human activity recognition (HAR) is the lack of large labeled datasets. This hinders the development of robust and generalizable models. 
Recently, cross modality transfer approaches have been explored that can alleviate the problem of data scarcity. 
These approaches convert existing datasets from a source modality, such as video, to a target modality (IMU). 
With the emergence of generative AI models such as large language models (LLMs) and text-driven motion synthesis models, language has become a promising source data modality as well as shown in proof of concepts such as IMUGPT. 
In this work, we conduct a large-scale evaluation of language-based cross modality transfer to determine their effectiveness for HAR. 
Based on this study, we introduce two new extensions for IMUGPT that enhance its use for practical HAR application scenarios: a motion filter capable of filtering out irrelevant motion sequences to ensure the relevance of the generated virtual IMU data, and 
a set of metrics that measure the diversity of the generated data facilitating the determination of when to stop generating virtual IMU data for both effective and efficient processing.
We demonstrate that our diversity metrics can reduce the effort needed for the generation of virtual IMU data by at least 50\%, which open up IMUGPT for practical use cases beyond a mere proof of concept.

%% file: figure/general_flow.tex


\begin{teaserfigure}
    \centering
    \begin{adjustbox}{width=0.75\linewidth, vspace=-0.3in}
        \includegraphics[scale=0.75]{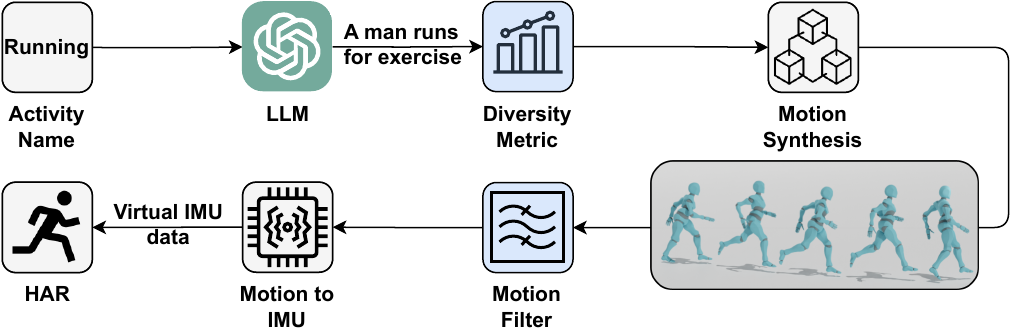}
    \end{adjustbox}
    \caption{Overview of the proposed language-based cross modality transfer system for sensor based human activity recognition. An LLM automatically generates textual descriptions of activities, which are then converted into motion sequences by a motion synthesis model.
    A novel motion filter then screens out incorrect sequences, retaining only relevant motion sequences for virtual IMU data extraction. 
    A new diversity metric measures shifts in the distribution of generated textual descriptions and motion sequences, allowing for the definition of a stopping criterion that controls when data generation should be stopped for most effective and efficient processing and best downstream activity recognition performance.
    }
    \label{fig:general_flow}
\end{teaserfigure}


%% file: sections/intro/main.tex
\section{Introduction}
The use of wearable sensing devices for Human Activity Recognition (HAR) is one of the central pillars of mobile and ubiquitous computing. 
HAR finds extensive application in various domains, including fitness tracking \cite{, chavarriaga2013opportunity, Koskim2017gym}, health monitoring \cite{bachlin2010wearable, Liaqat2019}, sign language recognition \cite{Liu2020gesture, Martin2023finger}, and in identifying critical events like vehicular accidents and falls \cite{ lahn2015car, liu2023fall}. 
The recognition is typically achieved through the use of supervised learning methods to classify segmented data streams into activities of interest (or the null class). 
The effectiveness of these methods relies heavily on the availability of accurately annotated datasets. 
However, one of the biggest challenges that the HAR community faces is the lack of large-scale labeled datasets due to the expensive annotation process, the need for domain experts, and potential issues with invasion of privacy 
\cite{kwon2019handling,jiang2021research,cilliers2020wearable, chen2021sensecollect, Plötz2023data}.

To address the problem of annotation scarcity, researchers have explored the use of cross modality transfer approaches. 
These approaches aim to convert data from other modalities, such as video \cite{kwon2020imutube}, motion capture \cite{xiao2021}, and optic data \cite{xu2022visual}, into \emph{virtual IMU data}. 
They leverage existing large-scale annotated datasets in the source data modality, and, in turn, use them to generate large-scale virtual IMU datasets with accompanying annotations.  
The generated data then serve as (additional) labeled training data for the downstream classifier to boost the model's performance. 

With the recent emergence of generative foundation models, natural language has become a new promising candidate data modality for cross modality transfer. 
Leng et al.\ \cite{leng2023generating} piloted the concept of IMUGPT that combines large language models (LLMs), motion synthesis methods, and signal processing techniques to generate virtual IMU data with no manual effort. 
The key idea is to use LLMs to generate diverse textual descriptions of the different ways that humans can perform certain activities. 
The generated textual descriptions are then converted to 3D human movement sequences using motion synthesis methods \cite{zhang2023generating}.
The resulting sequences of action-specific poses are then converted into virtual IMU training data. 
In principle, such a system allows for the generation of labeled datasets that are larger and encompass more activities than any existing ones. 
Such generated datasets would help pave the way for the development of more complex HAR models that are robust, generalizable, and allow for the analysis of more complex human movements and gestures, yet without involving any human participants. 
While the initial IMUGPT system served as a preliminary proof of concept, it demonstrated promise in small-scale experiments, where the generated virtual IMU data led to significant improvements in performance in the downstream classifier. 
This paper builds on those initial results by significantly  expanding on the proof of concept through a range of technical modifications and additions, that render the approach valuable for practical applications and by thoroughly evaluating it in a large scale experimental evaluation study.

Our goal is to determine not only \textit{how} but also \textit{how much} and \textit{what kind of} virtual IMU data shall be derived from the language-based input, for effective and efficient cross modality transfer.
We hypothesize that more diverse and relevant virtual IMU data leads to greater performance improvements for downstream applications. 
To ensure the relevance of the generated virtual IMU data, we introduce a motion filtering module capable of filtering out generated motion sequences that are irrelevant (due to limitations in motion synthesis models), preventing such virtual IMU data from being used for model training. 
Additionally, we also introduce two metrics to measure the diversity of virtual IMU data, through diversity of both textual descriptions and motion sequences. 
Subsequently, we correlate these metrics to the downstream classification performance, and also use them to determine the optimal stopping points for generating virtual IMU data.
Such stopping points indicate that the generated data have reached a saturation point in terms of diversity, beyond which, generating additional data does not enhance the diversity of the dataset nor benefit the performance of the downstream classifier. 
Through these extensions, IMUGPT can automatically determine the optimal quantity of virtual IMU data to generate. 
This provides researchers with efficient means to produce and utilize virtual IMU data, leading to more robust and accurate HAR applications while saving time and computing resources.

We further conduct a comprehensive evaluation of the IMUGPT system--in its extended form as presented in this paper--that explores the effectiveness of such a language-based cross modality transfer approach across a broader spectrum of activities over a range of practically relevant HAR application scenarios. 
This evaluation involves comparing different LLMs and motion synthesis models, with specific focus on  their impact on the performance of downstream activity recognition systems. 
Our results show that using GPT-3.5 \cite{openai2022gpt} and T2M-GPT \cite{zhang2023generating} for virtual IMU data generation lead to the best downstream performance. 
Additionally, we demonstrate that our proposed diversity metrics can reduce the time and compute resources needed for the data generation process by over 50\%, making IMUGPT more practically usable.


Through our methodological extensions and the extensive experimental evaluation of IMUGPT, we demonstrate the effectiveness of language-based cross modality transfer methods for deriving robust and effective HAR systems in practical application scenarios, thereby alleviating, if not circumventing, the most notorious challenge for contemporary HAR systems: the lack of large-scale, labeled training data.

%% file: sections/related_work/main.tex
\section{Related Work}

Sensor-based human activity recognition using wearable devices (HAR), i.e., the automated assessment of what a person is doing (and when), has come a long way and it is considered one of the central pillars of mobile, wearable, and ubiquitous computing \cite{kwon2021complex}.
Advances in the miniaturization of sensing hardware have enabled the widespread adoption and integration of inertial measurement units (IMUs) to capture movement information of a user as one of the most central contextual factors, that if analyzed properly, allows for situated and often personalized adaptation of computing services \cite{xiao2021}.
As such, a multitude of applications have been realized, which are based on the core activity recognition backend, including health monitoring \cite{bachlin2010wearable}, exercise and sports tracking \cite{Koskim2017gym}, and gesture recognition for intuitive user interfaces \cite{Liu2020gesture}, to name but a few.


Conventional HAR applications are typically based on (some variant of)  the "Activity Recognition Chain (ARC)" (e.g., \cite{bulling2014}), which divides the analysis problem into five general processing blocks: recording, preprocessing, (sliding-window) segmentation, feature extraction, and classification.
While the first two components are typically addressed through classical signal processing approaches, the remainder of the processing pipeline typically involves some form of machine learning methods, often following the supervised learning paradigm for which substantial amounts of \textit{labeled} training data are required.

More contemporary HAR workflows are based on end-to-end deep learning methods that promise more effective, generalizable models, for example, by overcoming the manual design of feature representations as it was required in the conventional ARC \cite{ploetz2011feature,plotz2018deep}.
Yet, even recent state-of-the-art end-to-end learning methods require substantial amounts of labeled sample data for model training -- an obstacle that cannot easily be overcome.

In what follows, we first give an overview of the most pressing challenges to sensor- and machine learning based human activity recognition with wearables. 
We then focus on related work in the field of general cross modality transfer as it is used to overcome shortages of labeled training data, which is the scope of the work presented in this paper.
We then give an overview specifically on language-based cross modality transfer, namely the original IMUGPT approach \cite{leng2023generating}, which serves as the basis for this paper.


\subsection{Challenges with HAR using Wearables}

Despite considerable advancements in HAR and its numerous practical applications, there exist significant challenges that are inherent and specific to HAR using wearable sensing platforms \cite{Plötz2023data, ravi2016, chen2021sensecollect, els2017}. The most notable ones are: (i) paucity of labeled data -- the time-consuming and expensive nature of wearable data collection along with its inherent privacy concerns have led to datasets being relatively smaller in size; (ii) difficulty in data annotation -- ambiguity in the target activities and its context results in incorrectly labeled data; (iii) conflicting variance in data -- induced by similar activities being performed differently and different activities resulting in similar sensor readings; and (iv) sensor noise -- auto-calibration of sensors that depend on temperature and gravity corrections \cite{vanhees2014env}, and the underlying architecture of MEMS sensors \cite{mohd2001mems} inducing noise in the data.

Arguably, the most pressing challenge is the lack of labeled training data. While recording data on wearable devices is uncomplicated due to constant activity tracking by sensors, the impracticality of either continuously recording a user's activities or asking users to repeatedly provide ground truth labels renders annotating the sensor data challenging, to say the least. The research community has developed a range of approaches to specifically tackle the sparse data problem, including self-supervised learning \cite{haresamudram2020masked, haresamudram2021cpc, tang2021selfhar, saeed2019ssl, haresamudram2022assessing}, few-shot learning \cite{FENG2019112782}, semi-supervised learning \cite{balabka2019semi}, prototypical learning \cite{bai2020proto, cheng2023protohar}, adversarial learning \cite{bai2020adversarial, leite2020adversarial}, and transfer learning \cite{sargano2017transfer}. Recently, the idea of cross modality transfer has gained traction in the community, in which, knowledge is transferred from one data modality to another, e.g., from image to text.

\subsection{Cross Modality Transfer}

cross modality transfer methods have recently been introduced in a number of application domains with the goal of opportunistically utilizing existing datasets from source modalities other than those targeted by a specific application.
The key motivation here is to combat the lack of domain and modality specific labeled data ("small labeled dataset problem") by automatically converting sensor readings from one modality to another, resulting in "virtual sensor data" in the target modality, thereby transferring knowledge in form of the underlying fundamentals of the human movements from one modality to another. 

 \input{tables/cross_modality_systems}

Approaches for generating virtual data through cross modality transfer from different data sources have been explored across various domains (\autoref{tab:cross_modality_systems}).  
cross modality transfer has been actively investigated for the generation of virtual IMU data for HAR from various sources. 
Xiao et al. \cite{xiao2021} use a Convolutional Neural Network (CNN) fine-tuned with real IMU data to generate "skinned multi-person linear" (SMPL) model  \cite{Loper2015smpl} parameters from motion capture data, which is further used to generate virtual acceleration and orientation data. 
An extension to this was presented by Uhlenberg et al.\ \cite{Uhlenberg2022mesh} who created 3D human surface models and skeletal models from motion capture data, which were then used to simulate daily activities and synthesize inertial data.

Xia et al. \cite{Xia2022} introduced a virtual spring-joint based sensor module to augment simulated virtual acceleration data extracted from 2D exercise videos. 
Recently, IMUTube \cite{kwon2020imutube, kwon2021approaching, kwon2021complex} was introduced to convert existing large-scale videos into virtual IMU data through a computer vision pipeline involving 2D pose extraction and conversion to 3D poses 
on which individual joints are tracked in order to generate tri-axial acceleration and gyroscope data. Similarly, Rey et al. \cite{rey2019} use a combination of real sensor data and 2D poses extracted from laboratory videos to train a regression model. The model takes arbitrary videos as inputs and generates simulated sensor data.

In the context of Sign Language Recognition, ZeroNet \cite{liu2021zeroshot} uses 3D finger pose extraction on publicly available sign language videos and compares it to IMU data retrieved from a finger ring from an unknown user gesture using a combination of Dynamic Time Warping and Convolutional Neural Networks. SignRing \cite{Li2023asl} approaches perform Sign Language Recognition using 
a triangulation-based algorithm to convert 2D videos of signers signing to 3D hand pose and then compute the sensor's 3D acceleration by tracking the movements of the index fingers. Vi2IMU \cite{Santhalingam2023asl} uses a combination of wrist and 3D displacement estimation with LSTM-based architectures based on 2D wrist, arm, and hand joint positions extracted from publicly available videos to generate virtual IMU data. 

Lu et al. \cite{lu2022} use Face Tracking \cite{baltrusaitis2018openface} and 6DRepNet \cite{Hempel_2022} to generate virtual IMU data for head motions from videos. VisualAcc \cite{xu2022visual} uses photometric effect-based interrogation and an optic-to-inertia transformer that senses human motion passively and reconstructs inertial data using the Optical Motion Field (OMF).



\subsection{Text-Driven Motion Generation}
A common challenge in the aforementioned cross modality transfer approaches is the gathering of relevant videos and motion capture datasets. 
Considering more niche applications like Sign Language Recognition, finding the right video dataset can be a time-consuming process. 
Furthermore, the quality of videos can be an issue in pose extraction and estimation. 
In order to address this, Leng et al. \cite{leng2023generating} proposed IMUPGT, a language-based cross-modal transfer approach that generates diverse virtual IMU data from virtual textual descriptions of activities using a combination of LLMs, motion synthesis models, and signal processing techniques, eliminating the need to search for videos. 
This system serves as the foundation for the work presented in this paper and more details about it are given in section \ref{sec:ref:background:imugpt}.

As a basis for approaches like IMUGPT, textual descriptions need to be converted into human motions that are performing underlying activities of interest.
Much research has gone into automatically generating 3D human motion using text descriptions of activities, which serves as a basis for the use of language as the source modality in cross modality transfer to generate virtual IMU data. 
With the recent introduction of the  HumanML3D dataset \cite{Guo_2022_CVPR} -- a large 3D human motion dataset with text descriptions -- motion synthesis and diffusion-based models like MotionGPT \cite{jiang2023motiongpt}, T2M-GPT \cite{zhang2023generating}, MotionDiffuse \cite{zhang2022motiondiffuse}, and ReMoDiffuse \cite{zhang2023remodiffuse} have been introduced that are capable of producing more realistic human motion sequences using textual descriptions as the input.

\textit{T2M-GPT and MotionGPT} use a motion tokenizer based on a Vector Quantized Variational Autoencoder (VQ-VAE) architecture, and a motion-aware language model based on a Generative Pre-trained Transformer (GPT) architecture. 
MotionDiffuse adopts a denoising diffusion probabilistic model (DDPM) using a cross modality linear transformer to convert a given text into a motion sequence. 
On the other hand, ReMoDiffuse enhances a denoising diffusion model using hybrid retrieval and uses a Semantic-Modulated Transformer for sequence generation. 
In our work, we explore and evaluate these motion synthesis models in the IMUGPT setup.


The emergence of \textit{Large Language Models} (LLMs) such as GPT-3 \cite{brown2020gpt3}, GPT-4 \cite{openai2023gpt4}, PaLM 2 \cite{anil2023palm}, and LLama 2 \cite{touvron2023llama} has revolutionized the field of natural language processing (NLP) due to their impressive capabilities in various NLP tasks, made possible by their massive training corpora. 
The success of LLMs has also attracted attention in other fields, where they are integrated with an array of AI models, with the LLMs facilitating the interaction between the AI models and the end-user \cite{shen2023hugginggpt, wu2023visual}. Additionally, in a recent study, Athanasiou et al. \cite{SINC:ICCV:2022} discovered that motion and activity information is encoded within LLMs, demonstrating that GPT-3 could be used to identify the body parts involved in different activities. Inspired by this result, we apply LLMs to the problem of activity filtering in this work to filter out irrelevant activity descriptions.

\subsection{IMUGPT}
\label{sec:ref:background:imugpt}
Pioneered by Leng et al.\ \cite{leng2023generating}, language-based cross modality transfer for sensor-based human activity recognition--IMUGPT (\autoref{fig:imugpt})--takes textual descriptions of relevant activities as input, which are then converted to sequences of 3D human motions, e.g., through the T2M-GPT model \cite{zhang2023generating}.
The IMUTube backend \cite{kwon2020imutube} is then applied to the resulting motion sequences and virtual IMU data are generated automatically that can then be used for training HAR models.
IMUGPT consists of three major components as follows:

\input{figure/imugpt}

\begin{enumerate}
\item \textbf{LLM}. \underline{What:} An LLM is used to generate diverse textual descriptions of a person performing a specified activity. \underline{Why:} Human activities are diverse in the real world, as the same activity can be performed in different ways by different people (e.g., a skinny teenager vs. a muscular athlete) and under different contexts (e.g., happily vs. sadly). To develop robust and generalizable HAR models capable of recognizing all these variations, it is essential to reflect this diversity in the training data. Generating diverse textual descriptions for activity performance is the first step toward ensuring that the generated virtual IMU data accurately represents real-world diversity. Additionally, using an LLM for text generation automates and simplifies the process, eliminating the need to search for video data as required in video-based cross modality transfer approaches.

\item \textbf{Motion Synthesis}. \underline{What:} A motion synthesis model that takes in textual descriptions of the activity and converts them into three-dimensional human motion sequences. \underline{Why:} The motion synthesis model bridges the gap between text and human motion. To extract virtual IMU data, understanding how a person moves in three-dimensional space is essential. This is similar to how three-dimensional poses are estimated in video-based methods.

\item \textbf{Motion to IMU}. \underline{What:} A method to convert generated motion sequences into virtual IMU data, either biomechanically \cite{Yound2011imusim} or through neural networks \cite{Santhalingam2023asl, Li2023asl}. \underline{Why:} The virtual IMU data extracted from the motion sequences can be used to train a deployable HAR model, either alone or in conjunction with real IMU data.
\end{enumerate}

IMUGPT was evaluated on locomotion activities \cite{Reisspamap, Sztyler2016realworld, Zhang2012usc}.  
Preliminary experiments demonstrated the promising potential of the approach, showing that the extracted virtual IMU data can lead to significant improvements in the downstream HAR performance.

%% file: tables/cross_modality_systems.tex
\begin{table}
\centering
    \caption{Overview of prominent cross modality transfer approaches for virtual data generation.} 
    \begin{adjustbox}{width=\columnwidth,center}
    \begin{tabular}{c|c|c|c|c}
        System & Source Data & Target Data & Task & Method   \\
        \hline
        \hline

        Liu  et. al. \cite{liu2019mri} & MRI & CT & Medical Diagnosis & Neural Network\\
        \hline

        Xia  et. al. \cite{xia2022using} & Distance & Image & Hand Gesture Recognition & Neural Network\\
        \hline
        
        Uhlenberg et. al. \cite{Uhlenberg2022mesh} & \multirow{2}{*}{Motion Capture} & \multirow{2}{*}{Gyro \& Acc} & \multirow{6}{*}{Human Activity Recognition} & Surface Modeling\\
        \cline{1-1}\cline{5-5}
        
        Xiao et. al. \cite{xiao2021} &  &  &  & \multirow{2}{*}{Neural Network}\\
        \cline{1-3}
        
        Zhang et. al. \cite{zhang2020cvae} & \multirow{10}{*}{Video} & \multirow{2}{*}{Acc} &  & \\
        \cline{1-1}\cline{5-5}
        
        Xia et. al. \cite{Xia2022} &  &  &  & \multirow{2}{*}{Pose Estimation \& Biomechanics}\\
        \cline{1-1}\cline{3-3}
         
        IMUTube \cite{kwon2020imutube, kwon2021approaching, kwon2021complex} &  & \multirow{2}{*}{Gyro \& Acc} &  & \\ 
        \cline{1-1}\cline{5-5}
        
        Rey et. al. \cite{rey2019} &  &  &  & \multirow{3}{*}{Pose Estimation \& Neural Network}\\
        \cline{1-1}\cline{3-4}
         
        ZeroNet \cite{liu2021zeroshot} &  & \multirow{2}{*}{Gyro \& Acc} & \multirow{3}{*}{Sign Language Recognition} & \\
        \cline{1-1}
         
        Vi2IMU \cite{Santhalingam2023asl} &  &  &  & \\
        \cline{1-1}\cline{3-3}\cline{5-5}
         
        SignRing \cite{Li2023asl} &  & Acc &  & Pose Estimation \& Biomechanics\\
        \cline{1-1}\cline{3-5}
         
        Lu et. al. \cite{lu2022}&  & Gyro & Head Motion Recognition & Pose Estimation \& Neural Network\\
        \cline{1-1}\cline{3-5}

        Vid2Doppler \cite{ahuja2021vid2doppler} &  & \multirow{2}{*}{Doppler} & \multirow{6}{*}{Human Activity Recognition} & \multirow{4}{*}{Neural Network}\\
        \cline{1-2}

        IMU2Doppler \cite{bhalla2021imu2doppler} & Gyro \& Acc &  &  & \\
        \cline{1-3}

        IMG2IMU \cite{yoon2022img2imu} & Image & \multirow{2}{*}{Gyro \& Acc} &  &  \\
        \cline{1-2}

        AudioIMU \cite{liang2022audioimu} & Audio &  &  & \\
        \cline{1-3}\cline{5-5}

        Visual Accelerometer \cite{xu2022visual} & Optic & Acc &  & Pose Estimation \& Biomechanics \\
        \cline{1-3}\cline{5-5}
         
        Leng et. al. \cite{leng2023generating} & Text & Gyro \& Acc &  & Motion Synthesis \& Biomechanics\\
       \hline
        
    \end{tabular}
    \end{adjustbox}
    \label{tab:cross_modality_systems}
    
\end{table}

%% file: figure/imugpt.tex


\begin{figure}[t]
    \centering
    \begin{adjustbox}{width=1\linewidth, vspace=-0.3in}
        \includegraphics[scale=0.75]{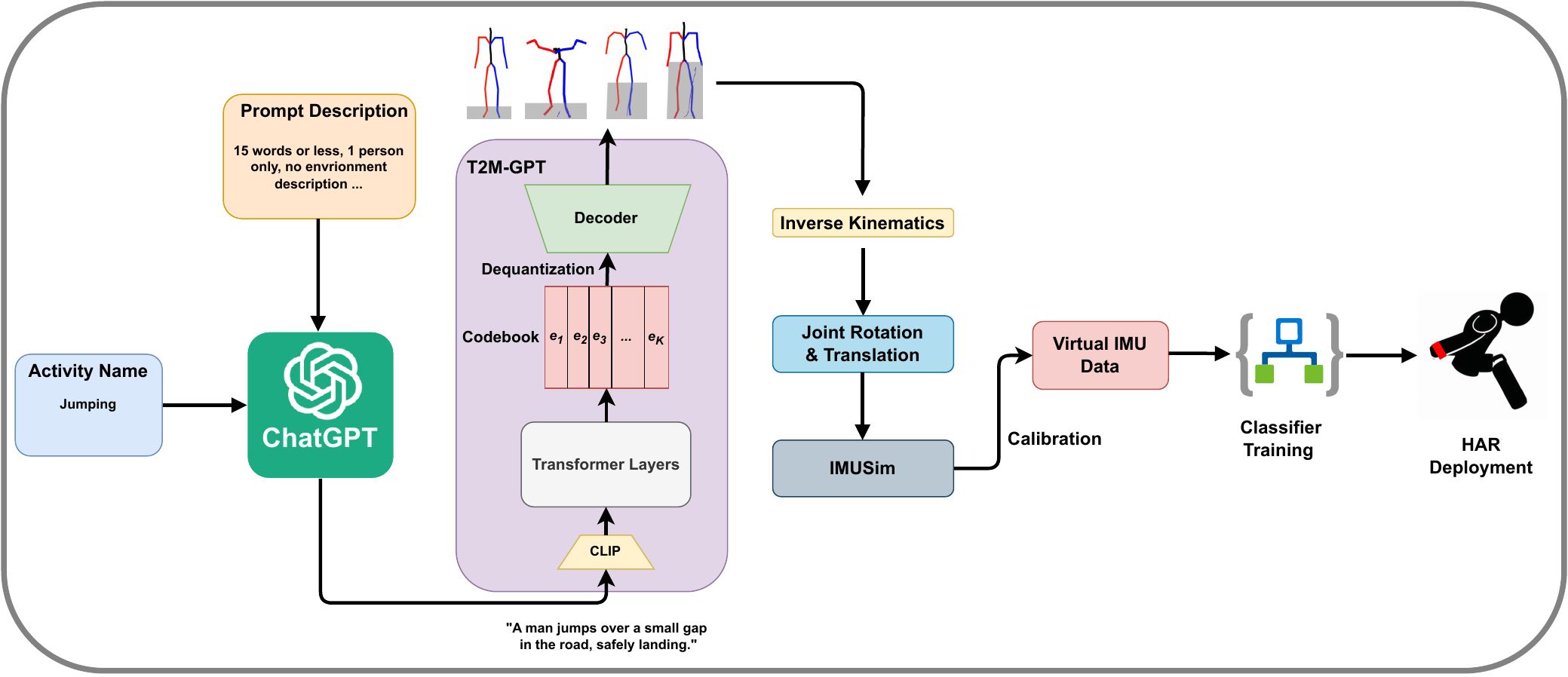}
    \end{adjustbox}
    \caption{Overview of Leng et al.'s IMUGPT \cite{leng2023generating}. ChatGPT is used to generate diverse textual descriptions of the specified activities. Subsequently, a motion synthesis model, T2M-GPT \cite{zhang2023generating}, generates human motion sequences using the textual descriptions. Virtual IMU data can then be extracted from the generated motion sequences and used for training HAR models.    (Figure adopted from Leng et al. \cite{leng2023generating} and used with permission)
    }
    \label{fig:imugpt}
\end{figure}


%% file: sections/method/main.tex
\section{IMUGPT 2.0: Towards Practical Applications of Language-Based Cross Modality Transfer for HAR}

\input{figure/general_flow}
While the initial prototype of IMUGPT \cite{leng2023generating}, as summarized above, demonstrated the general effectiveness of the idea of language-based cross modality transfer in sensor-based human activity recognition, that proof of concept fell short on two major aspects:
\textit{i)} It is unclear when data generation should be stopped, which has implications on the computational efficiency (and costs) of the virtual IMU data generation process; and
\textit{ii)} The relevance of the generated virtual IMU data remains uncertain until a downstream HAR system has been trained and evaluated. It may very well be the case that some generated IMU data are not useful or are possibly even detrimental to the targeted HAR application scenario.

In this work, we present extensions to the initial IMUGPT prototype through which we add two additional modules thereby explicitly aiming to address aforementioned limitations.
\autoref{fig:general_flow} illustrates these extensions in blue. 
These new modules facilitate language-based cross modality transfer, making it more practical as follows:

\begin{enumerate}
\item \textbf{Diversity Metrics}. \underline{What:} A method to measure the diversity of textual descriptions and motion sequences generated by LLMs and motion synthesis models, respectively. See section \ref{sec:diversity} for details. \underline{Why:} We hypothesize that the downstream model performance depends on the diversity of the virtual IMU data, as more diverse data can enhance model performance. With the proposed diversity metrics, we introduce a saturation point identification algorithm (\autoref{alg:sat_point_ind}) 
that identifies the point at which generating additional textual descriptions no longer adds meaningful information to the existing pool of generated texts. This indicates a stopping point for text generation. Automatically identifying when to stop data generation saves time and computing resources, which are not unlimited.

\item \textbf{Motion Filter}. \underline{What:} A pipeline that identifies and filters out motion sequences that do not accurately portray the specified activity. See section \ref{sec: motion_filter} for details. \underline{Why:} The motion synthesis model does not always generate motion sequences that accurately portray a person performing the specified activity, as it can confuse closely related activities (e.g., climbing up stairs vs. climbing downstairs), resulting in irrelevant motion sequences. We hypothesize that irrelevant motion sequences will negatively impact the downstream classifier's performance by introducing noise. 
\end{enumerate}




\subsection{Diversity Metrics}
\label{sec:diversity}
A key limitation of IMUGPT \cite{leng2023generating} is that there is no clear indication as to when the generation of data should be stopped. While this was acceptable for a proof of concept, in order to make the system viable and practical, it is important to have an explicit stopping point for data generation. This would ensure control over the time and computation costs associated with the generation process, as well as making sure that the generated virtual data actually are of benefit for the HAR modeling task. For this reason, we propose utilizing the diversity of the generated data to determine the optimal stopping point. Note that the diversity of the virtual IMU data can be measured either through the textual descriptions or the motion sequences it is generated from. Since the motion sequences are obtained from the textual descriptions themselves, we hypothesize that the diversity of both should be correlated. We validate this in section \ref{sec: diversity_predictor}. Intuitively, calculating the diversity of the textual descriptions is more practical as it precedes the motion sequence generation step, saving computational costs. 

Diversity enables us to quantify the amount of useful information present in the data, and then utilize that to make a judgment about downstream performance without running the entire pipeline. 
If the diversity is high, then downstream models would be exposed to a broader range of data points during training. This results in the model learning to make predictions on a wider variety of cases, thereby improving the performance. Diverse training data also helps prevent the models from overfitting, since a limited dataset would lead to overexposure to a small section of the feature space. It is helpful to have some practical measures that facilitate the estimation of downstream performance at the time of generation itself, assuming that diversity is computationally inexpensive to calculate. 


In order to calculate the diversity of data (whether textual descriptions or motion sequences), we generate embeddings for the data. Embeddings are vectorial representations of data that serve to capture the semantic and syntactic information present in the data. Each data point is mapped to a vector in an `embedding space'.
Since vector operations can be utilized to operate on the data in an embedding format, they are a useful representation for calculating diversity metrics.
We generate embeddings as follows: 
\begin{itemize} 
    \item \textbf{Textual descriptions}: The text prompts are passed through SentenceTransformers's `all-mpnet-base-v2 model' \cite{reimers-2019-sentence-bert, all-mpnet-base-v22024} 
    to generate embeddings for each prompt. This model was trained on one billion sentence pairs to capture the semantic information of its input text, and thus, the generated embeddings serve as a suitable representation of the sentence. 
    
    \item \textbf{Motion Sequences}: Each motion sequence is passed through a model trained on the HumanML3D dataset 
    \cite{Guo_2022_CVPR} to generate the embeddings for the sequence. This model is drawn from the motion feature extractor trained in Guo et al.\ \cite{Guo_2022_CVPR}, and is commonly used in the community \cite{zhang2023generating}.
\end{itemize}

Overall, we compute two types of diversity:
\textit{i)} absolute diversity, and 
\textit{ii)} comparative diversity. 
Note that these metrics are applicable to any sets of embeddings, irrespective of whether they are generated from textual descriptions or motion sequences.
The diversity calculation methods themselves are universally applicable to any collection of embeddings. Absolute diversity provides a quantitative measure for the amount of diverse information contained in a set, while comparative diversity showcases how much two sets differ in terms of their diversity. These measures are detailed further in the subsequent sections. 
\subsubsection{Absolute Diversity Metrics}\label{sec:abs_div_metric}
We use two methods to compute the absolute diversity of a set of embeddings -- the standard deviation method \cite{lai_diversity_nodate} and the centroid method \cite{Duda2000}. For both metrics, a higher value corresponds to a more diverse set, and vice versa. 

\paragraph{Standard Deviation Metric}
The standard deviation method is based on the diversity metric introduced in Lai et al.\ \cite{lai_diversity_nodate}. Interpreting embeddings as vectors in a high-dimensional embedding space, the goal is to characterize the dispersion (the "spread") of a cluster of such vectors. 
If the cluster is assumed to be distributed as a multi-variate Gaussian, each isocontour will be shaped as an axis-aligned ellipsoid. 
The radii of the ellipsoid along each axis can be computed by calculating the standard deviation 
of the vectors in the cluster along each of the axes. 
Thus, computing the geometric mean of the radii will capture the generalized radius of the cluster, providing a metric for the diversity. 
Assuming the set of embeddings $S$ consists of $n$ embedding vectors, each of dimension $k$, the set can be formalized as $ S = \left\{ x_i \right\}_{i=1}^n \in \mathbb{R}^k$. 
Thus, the standard deviation along an axis $ j \in [1, k] $ is computed as follows:

    \begin{equation}
        \sigma_j = \sqrt{\frac{\sum_{i=1}^{n}\left(x_i^j - \mu^j\right)^2}{n}} \text{, wherein }  \mu^j = \frac{\sum_{i=1}^{n}x_i^j}{n}  
    \end{equation}
    The final diversity score $M_{std}$ is then calculated as:
    \begin{equation}
        M_{std} = \left( \prod_{j=1}^{k} \sigma_j \right)^{\frac{1}{k}}
    \end{equation}

\paragraph{Centroid Metric \cite{Duda2000}}
Assume the set of embeddings $ S = \left\{ x_i \right\}_{i=1}^n \subset \mathbb{R}^k$, consisting of $n$ embedding vectors, each of dimension $k$. First, the centroid vector $x_{cent}$ of all  embedding vectors is calculated: 
\begin{equation}
    x_{cent} = \frac{\sum_{i=1}^{n}x_i}{n}  
\end{equation}
The diversity score $M_{cent}$ is computed as the mean of the sum-squared distances of all  embedding vectors from the centroid vector:
\begin{equation}
    M_{cent} = \frac{\sum_{i=1}^{n}\left(d(x_i, x_{cent})\right)^2}{n} \text{, wherein $d$ is the Euclidean distance }  d(x_i, x_{cent}) = \sqrt{\sum_{j=1}^{k}\left(x_i^j - x_{cent}^j\right)^2}
\end{equation}
The intuition for this method stems from the fact that the higher the diversity for a set of embeddings, the farther apart their vectors will be spread in the embedding space. As a result, the points will tend to be at a farther distance from the overall mean than for a less diverse set, and the centroid metric quantifies this aspect. Since our goal is to quantify the dispersion of all the vectors in the embedding space, we only utilize a single centroid for the entire set of embeddings.

\input{figure/algorithm}

\subsubsection{Comparative Diversity}\label{sec:comparative_diversity}
The textual description generation process starts by generating an initial set of descriptions. Then, newly generated batches of textual descriptions are appended sequentially to the pre-existing set of descriptions. 
If adding a new batch of textual descriptions improves the diversity of the pre-existing set, then continuing to further generate newer batches is well-motivated as it would contribute to the diversity of the virtual IMU data, leading to better downstream performance. On the contrary, if diversity does not improve, newer batches should not be generated. 
The goal of comparative diversity is to quantify this change in diversity that occurs upon adding a new batch to the pre-existing set. 

    

The comparative diversity between two sets of embeddings is computed using the Maximum Mean Discrepancy (MMD) test \cite{JMLR:v13:gretton12a}. MMD is a kernel-based statistical test which indicates whether two given sets of samples are drawn from the same distribution or not. 
The higher the value, the farther the distributions of the two sets of samples. Given a space \(\mathbb{R}^d\) and independent and identically distributed samples \(X_i \in \mathbb{R}^d, i = 1, \ldots, N_x\) sampled from \(X \sim P_X\)
and \(Y_i \in \mathbb{R}^d, i = 1, \ldots, N_y\) sampled from \(Y \sim P_Y\), the MMD quantifies the difference between \(P_X\) and \(P_Y\). It is calculated as follows: 
\begin{equation}
MMD = \sum_{i=1}^{N_x} \sum_{j=1}^{N_x} K(X_i, X_j) + \sum_{i=1}^{N_y} \sum_{j=1}^{N_y} K(Y_i, Y_j) - 2 \sum_{i=1}^{N_x} \sum_{j=1}^{N_y} K(X_i, Y_j).
\end{equation}
Here, $K$ is the Gaussian kernel. Thus, we interpret any two sets of embedding as being drawn from two distributions, and calculate the ``distance'' between these two distributions to serve as the difference in diversity. 
If the distributions of the two sets of embeddings are similar, they will have a similar diversity, and correspondingly the MMD value for the pair of sets will be lower. 
Further details about how the comparative diversity is utilized to halt the generation process is highlighted in the subsequent  sections.

\subsubsection{Saturation Point Identification}
\label{sec: algorithm}

Given the comparative diversity measure introduced in the previous section, our goal is now to utilize this metric in a manner wherein the generation of textual descriptions can be halted once the generated set of descriptions saturates (i.e. generating further textual descriptions would not improve the diversity of the pre-existing set).
In order to determine this halting point, we propose the Saturation Point Identification algorithm  (\autoref{alg:sat_point_ind}). 
Given a pre-existing set of textual descriptions $S$ of size $n$, the algorithm iteratively generates new batches of descriptions and tests if adding the new batch to the pre-existing set alters the comparative diversity significantly. 
The size of the new batch is a percentage of the size of the pre-existing set, and is a tunable hyperparameter. 
Setting a higher value for the percentage would imply a coarser steps towards the saturation point. 

If adding the new batch alters the diversity substantially, then the diversity hasn't saturated yet, and the iteration continues. On the other hand if the diversity does not change meaningfully, it may be close to reaching saturation, and the algorithm terminates.
Actual termination is controlled via a hyperparameter that controls for tolerance for saturation. 
Note that for calculating the MMD between two sets of embeddings, the sets must have the same size. However, in this case, we calculate the MMD between two sets of differing sizes (in \autoref{alg:sat_point_ind}, these are $S\_emb$ and $S\_emb \cup B\_emb$ respectively). In order to address these, we randomly resample the smaller set (since random resampling does not alter the underlying population distribution
) until it has the same size as the larger set and then calculate the MMD. Note that this process of resampling is conducted multiple times, resulting in multiple values of MMD, and thus the function returns the mean score along with the standard deviation for MMD calculation ($MMD\_calculator$ function in  \autoref{alg:sat_point_ind}). Our algorithm is evaluated in section \ref{sec: algo_eval}. 

\subsection{Filtering out incorrectly generated motion sequences}
\label{sec: motion_filter}

\input{figure/motion_filter}

Another limitation of IMUGPT is that the motion synthesis model may generate motion sequences that do not accurately describe the intended activity, which leads to irrelevant virtual IMU data being extracted, potentially degrading the downstream performance. To address this, we propose a  motion filter that can filter out incorrectly generated motion sequences.
\autoref{fig:motion_filter_fig} illustrates the operation of the motion filter. To determine if a given motion sequence accurately portrays the activity of interest, the sequence generated by the motion synthesis model is first processed by a motion captioning model \cite{jiang2023motiongpt}. This model outputs a textual description of the motion sequence. The resulting motion caption is then evaluated by an LLM, which provides a binary `yes' or `no' answer indicating whether the caption accurately describes the specified activity. This output from the LLM enables us to filter out any incorrectly generated motion sequences.


\subsubsection{Motion Captioning}
The motion captioning task, also called the motion-to-text task, refers to generating a text description for a given human motion sequence. TM2T \cite{Guo2022tm2t} and MotionGPT \cite{jiang2023motiongpt} are two recently introduced models for motion captioning. 
In this work, we use MotionGPT to caption the generated motion sequences due to its superior performance. 
MotionGPT is a motion-language model that is trained on large amounts of language data and motion data to handle numerous motion-relevent tasks such as text-driven motion
synthesis, motion captioning, motion prediction, and motion in-betweening. 

To caption a given input motion sequence of length $M$ frames $m^{1:M} = \{x_i\}_{i=1}^M$, the motion sequence is first encoded into $L$ discrete motion tokens $z^{1:L} = \bm{\varepsilon}(m^{1:M}) = \{z_i\}_{i=1}^L, L = M/l$, 
using a motion encoder $\bm{\varepsilon}$, where $l$ is the temporal downsampling rate on motion length. The motion encoder is part of a motion tokenizer that is based on the Vector Quantized Variational Autoencoders (VQ-VAE) architecture \cite{oord2017vq-vae}. This allows the motion sequence to be represented as a language. The motion encoder consists of two parts: 1D convolutions and quantization. In the first part, 1D convolution layers are applied to the motion sequence $m^{1:M}$ along the time dimension to obtain the latent vectors $\hat{z}_{1:L}$. Through discrete quantization, the latent vectors are converted into code indices using a learnable codebook  $Z = \left\{ z_i \right\}_{i=1}^K \subset \mathbb{R}^d$ with $K$ latent embedding vectors of dimension $d$. During quantization, each latent vector is replaced with the nearest vector in $Z$, which minimizes the Euclidean distance:
\begin{equation}
z_i=\arg\min_{z_k\in Z}\left\|\hat{z}_i-z_k\right\|_2
\end{equation}
Here, $z_i$ is the quantized latent vector.
The code indices corresponding to the quantized latent vectors in the codebook are called motion tokens, which can be interpreted as the vocabulary for human motion.

In addition to taking in a motion sequence as input, MotionGPT also processes a textual prompt describing the specific task to be performed, in this case, motion captioning. 
Similar to the input motion sequence, the input textual prompt is tokenized using the text encoder of a pre-trained text tokenizer, SentencePiece \cite{kudo2018sentencepiece}, with a vocabulary of $K_t$ text tokens. 
The text vocabulary $V_t = \{v_t^i\}_{i=1}^{K_t}$ is combined with the motion vocabulary $V_m = \{v_{m}^i\}_{i=1}^{K_m}$, which includes motion tokens and additional special tokens (e.g., the tokens indicating the start and end of a motion sequence) to form the unified vocabulary $V = \{V_t, V_m\}$ of a language model. 
The tokens/words within the new vocabulary can represent text, human motion, or a mixture of the two. 
This flexibility in representation allows MotionGPT to use both text and motion as input and output to accomplish a range of motion-related tasks with a single model.

The sequence of text and motion tokens $X_s = \{x_s^i\}_{i=1}^N
$, $x_s \in V$, are passed to a transformer-based language model as input. The language model generates a sequence of output tokens $X_t = \{x_t^i\}_{i=1}^L
$, in an autoregressive manner, and it is trained using the following loss function

\begin{equation}
\mathcal{L}_{LM} = -\sum_{i=0}^{L_t-1} \log p_\theta (x_t^i | x_t^0, \ldots , x_t^{i-1}, x_s)
\end{equation}
Lastly, the sequence of output tokens is decoded using a text decoder and a motion decoder to recover the output texts, motion caption in this case. 

In our implementation, the codebook size of the motion encoder is set to $Z \in \mathbb{R} ^{512x512}$. 
Additionally, the motion encoder uses a temporal downsampling rate of 4. 
T5 \cite{Colin2020t5} is chosen as the transformer-based language model with 12 layers in both the encoder and decoder. 
The model is trained on the HumanML3D dataset \cite{Guo_2022_CVPR}, a dataset containing large amounts of human motion capture data along with the corresponding textual descriptions, using the AdamW optimizer. 
We use the pre-trained model that Jiang et al.\ \cite{jiang2023motiongpt} released.\footnote{https://motion-gpt.github.io}

\subsubsection{Activity Filtering Using Large Language Models}
\label{sec:llm_filter}


The problem is framed as a binary classification task where the input is a textual description $m$, the motion caption of a human motion sequence, along with a specific activity of interest $a$. The output is a binary label $y \in \{ 0,1\}$ that indicates whether the input textual description accurately depicts a person performing the activity in question. Let $\mathcal{M}$ represent the space of all possible textual motion captions, and let $\mathcal{A}$ represent the set of all predefined activities of interest. We seek to learn a function $f$ that maps a pair consisting of a motion caption and an activity to a binary label:  $ f: \mathcal{M} \times \mathcal{A} \rightarrow \{0, 1\}$. The goal of $f$ is to determine whether a given motion caption $m \in \mathcal{M}$ correctly describes a person performing a given activity $a \in \mathcal{A}$. With conventional supervised methods in NLP, learning such a function would require large amounts of training data containing motion captions for specific activities. However, collecting such a dataset would be extremely time-consuming and practically infeasible. 


Hence, we decided to turn to LLMs for this problem as they have shown impressive performances on zero-shot NLP tasks \cite{brown2020gpt3, kojima2023large}. Through our early explorations, we believe the function $f$ has been implicitly learned by the LLMs during the training process due to the massive training corpus, as we empirically found that the LLMs have a good understanding of the correspondence between the motion descriptions and activities.

To obtain labels from the LLM, we first assign it a task via a system message, which is a type of message used to define the role of the LLM. In this message, the LLM is asked to provide `yes' or `no' responses, indicating whether the user-provided motion captions describe the specified activity accurately. Using this system message helps the LLM understand its role and significantly reduces post-processing effort. Without this message, we found that the LLM often produces additional, miscellaneous texts unrelated to the task. After setting up the system message, we provide the specified activity name and a list of motion captions, with 10 captions per prompt, to the LLM. The LLM then outputs `yes' or `no' for each caption. A `yes' response indicates that the caption correctly describes someone performing the specified activity, and `no' indicates otherwise. We use the LLM's responses to filter out incorrectly generated motion sequences. 

This approach is a zero-shot task for the LLM, as we do not provide any example captions and labels for it to learn from. We detail the exact prompts used, along with some example motion captions and the corresponding labels generated by the LLM, in \autoref{fig:motion_filter_llm} in \autoref{app:motion_filter_LLM}. For the labels shown in \autoref{fig:motion_filter_llm}, we used GPT-4 \cite{openai2023gpt4} as our LLM. It's important to note that the LLM is not infallible; for example, it incorrectly labeled the caption 'a person jogs in place then stops' as `no' for the running activity.  We evaluate how well the proposed motion filter can filter out incorrectly generated motion sequences in section \ref{sec: motion_filter_eval}.

%% file: figure/algorithm.tex
\SetKwComment{Comment}{// }{}
\begin{algorithm}[t]
\small
\DontPrintSemicolon
\caption{Saturation Point Identification 
}\label{alg:sat_point_ind}
\textbf{Function to calculate MMD:} $MMD\_calculator$ \;
\textbf{Pre-existing set of text embeddings:} $S\_emb[n]$ \Comment*[r]{n: size of the set}

$perc \gets 0.05$ \Comment*[r]{Percentage of text prompts generated at each step}
$early\_stop \gets 5$ \Comment*[r]{Hyperparameter to determine the stopping point}
$stop\_count = 0$ \;
$range\_min, range\_max \gets -1, -1$ \;

\While{$stop\_condition \leq early\_stop$}{
    $B[perc*n] \gets$ Generate new batch of prompts \;
    $B\_emb[perc*n] \gets$ Compute embeddings for B \;
    $score, standard\_dev \gets MMD\_calculator(S\_emb, S\_emb + B\_emb) $ \;
    \eIf{$ score > range\_min$ and $ score < range\_max$}{
      $stop\_condition += 1$\;
      $range\_min = min(score - standard\_dev, range\_min)$ \;
      $range\_max = max(score + standard\_dev, range\_max)$ \;
    }  
    {
      $range\_min = score - standard\_dev$ \;
      $range\_max = score + standard\_dev$ \;
      $stop\_condition = 0$\;
    }  
    $S\_emb \gets S\_emb \cup B\_emb$  \Comment*[r]{Append current batch to the pre-existing set}
}
return $S\_emb$
\end{algorithm}


%% file: figure/motion_filter.tex
\begin{figure*}[t]
    \centering
    \includegraphics[width=\linewidth]{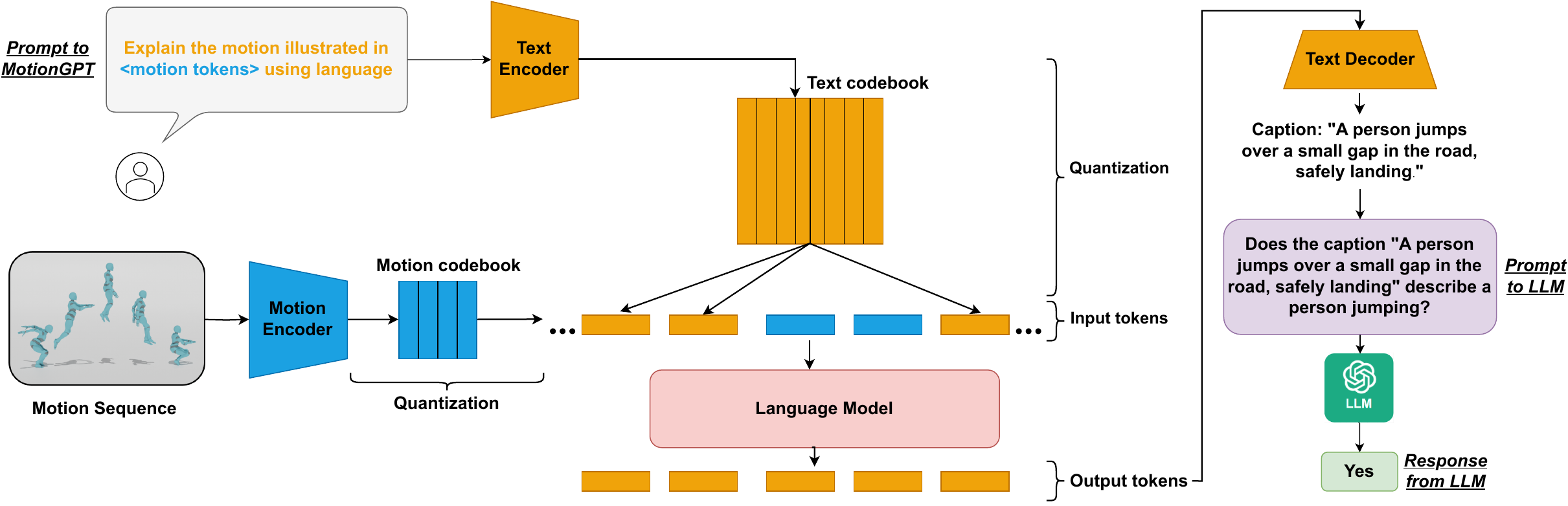}
    \vspace{-0.1in}
    \caption{Overview of the proposed motion filter. Using MotionGPT \cite{jiang2023motiongpt}, we obtain motion captions which are textual descriptions of the input motion sequence. To obtain the motion caption, a language model takes in encoded text and motion tokens and then generate output tokens. The output tokens are decoded to recover the motion caption. Then, we pass the motion caption into a LLM to determine if the motion sequence correctly portrays a specified activity. 
    }
    
    \label{fig:motion_filter_fig}
\end{figure*}

%% file: sections/experiment/main.tex
\section{On the Effectiveness of Language-based Cross Modality Transfer for HAR}

\label{sec:experiment}

As the second major contribution, we conduct 
an extensive study on the--now extended--language-based cross modality transfer approach: we run a large-scale experimental evaluation of IMUGPT 2.0 that specifically focuses on the practical aspects of this new cross modality transfer paradigm with the goal of assessing its relevance for real-world HAR applications. 
We run our study in two parts. 
The first part evaluates the original IMUGPT, i.e., the first language-based cross modality transfer approach, without our newly proposed components.
In contrast to the pilot study conducted by Leng et al.\ \cite{leng2023generating}, this experimental evaluation is on a much larger scale with larger numbers of activities compared to the previous study \cite{leng2023generating}. 
We conduct experiments using different LLMs (section \ref{sec: LLM_eval}) and motion synthesis models (section \ref{sec: motion_eval}) for generating textual descriptions and motion sequences. 
Through these experiments, we examine the impact that different generative models may have on downstream HAR performance and, in turn, reveal insights into the suitability of models, which will help practitioners design language-based cross modality transfer systems. 

In the second part of our study, we conduct an evaluation using our newly proposed components, i.e., the motion filter and diversity metric, thereby building on the results of part 1 of our study.
We start with the proposed diversity metrics and validate our hypothesis that diversity in the generated textual descriptions and motion sequences can serve as a predictor for downstream model performance and evaluate the effectiveness of our proposed saturation point identification algorithm (section \ref{sec: diversity_predictor}). 
Following this, we evaluate the new motion filter and determine how effectively it can filter out incorrectly generated motion sequences and its impact on the downstream model performance (section \ref{sec: motion_filter_eval}).

\input{sections/experiment/dataset_model}

\input{sections/experiment/llm_eval}

\input{sections/experiment/motion_eval}

\input{sections/experiment/diversity}

\input{sections/experiment/motion_filter}

%% file: sections/experiment/dataset_model.tex
\subsection{Human Activity Recognition}

\subsubsection{Datasets}

In line with previous work in the field, we conduct our evaluation on five public HAR datasets:
\textit{i)} RealWorld \cite{Sztyler2016realworld};
\textit{ii)} PAMAP2 \cite{Reisspamap};
\textit{iii)} USC-HAD \cite{Zhang2012usc};
\textit{iv)} HAD-AW \cite{Mohammed2018hadaw}; and 
\textit{v)} MyoGym \cite{Koskim2017gym}. 
These datasets contain IMU data recorded from varying on-body locations like head, chest, arm, waist, and leg for daily activities. 
Each dataset covers different activities and a varying number of subjects as summarized in \autoref{tab:dataset} in \autoref{app:datasets}.

\subsubsection{Classification Model}
Our evaluations are based on three models that are widely used in the HAR research community: 
\textit{i)} a Random Forest classifier;
\textit{ii)} DeepConvLSTM \cite{convlstm}; and 
\textit{iii)} DeepConvLSTM with self-attention \cite{singh2021attention}. 
We use sliding windows two seconds long with a 50\% overlap between consecutive frames to split the IMU data. 
We trained the Random Forest classifier on ECDF features \cite{hammerla2013preserving} (15 components). 
We trained DeepConvLSTM and DeepConvLSTM with self-attention on IMU raw data for a maximum of 30 epochs with an Adam optimizer and a ReduceLROnPlateau learning rate scheduler \cite{reducelronplateau}. 
We used grid search to determine the learning rate and weight decay. 
The learning rate varied from $10^{-6}$ to $10^{-2}$ and the weight decay varied from $10^{-4}$ to $10^{-3}$. 
For RealWorld, PAMAP2, USC-HAD, and MyoGym datasets, we used leave-one-subject-out cross validation for all 3 models. 
For the HAD-AW dataset, we used 5-fold stratified cross-validation instead since not all subjects performed the entire set of activities in the released dataset. 
For each model, we repeat the cross-validation for 3 random seeds and report the average macro F1 score and the standard deviation across the three runs.

%% file: sections/experiment/llm_eval.tex
\subsection{Impact of LLM on Textual Description Generation and HAR}
\label{sec: LLM_eval}

We explored various LLMs to explore their capability to generate textual descriptions regarding diverse human activities, and their effect on downstream classifier performance. 

\subsubsection{Experimental Setting}

We generated motion descriptions using five LLMs:
\textit{i)} GPT-3.5 \cite{openai2022gpt};
\textit{ii)} GPT-4 \cite{openai2023gpt4};
\textit{iii)} Palm 2 (Bard) \cite{anil2023palm};
\textit{iv)} Gemini \cite{geminiteam2023gemini}; and 
\textit{v)} LLaMa 2 \cite{touvron2023llama}. 
All LLMs, except LLaMa 2, were available as APIs, while the LLaMa 2 model with 70 billion parameters was deployed and hosted on a server for use as an API. 
An automated pipeline was developed with parameters for the type of model and the dataset. 
We briefly experimented with each LLM to determine the optimal prompt for the generation of text descriptions. 
The activities from all five datasets were provided to the LLMs as a text, along with sample descriptions and the LLMs were asked to describe a person performing the activity.

The generation of 1,000 descriptions for each activity was done in batches of 50, as descriptions for larger batches resulted in errors. 
The resulting responses were parsed programmatically to remove empty lines, serial numbers, and responses containing auxiliary text like ``Here are the descriptions for the activity''. 
The cleaned descriptions were passed into the motion synthesis models to create motion sequences, and the virtual IMU data was extracted, following the IMUGPT pipeline. The sizes of the real IMU dataset and the virtual IMU datsets are shown in \autoref{tab:llm_data_size} of \autoref{app:dataset_size}. 


\subsubsection{Qualitative Evaluation of Generated Text}
\label{subsubsec:qual_eval}
Examples of the generated textual descriptions are listed in \autoref{tab:more_prompts} in \autoref{app:text_descriptions}. We also provide all generated textual descriptions in the supplementary material for this paper. 
The generated textual descriptions were diverse on first look but an analysis revealed interesting results. 
On programmatic evaluation, GPT-3.5, GPT-4, and LLaMa 2 generated descriptions for the activities with minimal errors. 
Examples of errors include:
\textit{i)} inability to generate descriptions -- the LLM responds with ``Sorry, I could not generate descriptions for the activity'';
\textit{ii)} incorrect activities described -- the LLM generate text descriptions that did not describe the activity, instead described an emotion associated with the activity. 
On the contrary, Palm 2 and Gemini had conflicting results. 
In some instances, the text described inanimate objects (sponge, rocket) and animals (cat, dog) performing the activities. 
In other cases, the models generated highly repetitive descriptions for a particular activity. 
The generation of prompts with Palm 2  and Gemini were also tedious with a larger number of failures, irrelevant text, and missing text. 
This was primarily noticeable in complex activities like workout-related activities present in MyoGym. Overall, GPT-3.5, GPT-4, and LLaMa 2 produced better textual descriptions of activities and expect the generated virtual IMU data to lead to better downstream performance compared to Palm 2 and Gemini. 


\input{tables/llm_eval}

\subsubsection{Results} 

\autoref{tab:llm_eval} shows the downstream performance when different LLMs are used in IMUGPT for virtual IMU data generation. Overall, we find that the downstream performance is best when GPT-3.5 is used to generate the textual descriptions. With the Random Forest classifier, the downstream performance of GPT-3.5 shows an overall improvement of 1.0\% over GPT-4, 0.9\% over LLaMa 2, 0.6\% over Palm 2 , and 0.8\% over Gemini across all the datasets.
Therefore, we will use GPT-3.5 for generating textual descriptions in subsequent experiments unless otherwise specified.
We were surprised that Gemini and Palm 2  achieved competitive
downstream performance (less than 1\% difference compared to the other three LLMs), despite them producing qualitatively worse textual descriptions with less context information (as discussed in section \ref{subsubsec:qual_eval}). We discuss the effect of additional context information within the textual descriptions in section \ref{sec: prompt_enginnering}.

When using the Random Forest classifier, the generated virtual IMU data led to significantly better downstream performance for all datasets but one, HAD-AW. For the deep learning models, Deep ConvLSTM and Deep ConvLSTM with self-attention, the results were mixed. The virtual IMU data did not significantly improve model performance for USC-HAD, HAD-AW, and the MyoGym datasets. We attribute this to the deep learning models overfitting to the virtual IMU data, where the models learned features only applicable to the virtual IMU, hence, the deep learning models were unable to generalize to the testing dataset.



%% file: tables/llm_eval.tex
\begin{table}[t]
  \centering
  \caption{Model performance (Macro F1 score) when trained on both the real IMU data and the virtual IMU data when various LLMs are used to generate the textual descriptions in IMUGPT.  ``Real Data'' denotes the baseline experiments not including any generated, virtual IMU data. 
  }
    \small
  \begin{tabular}{l c c c c c}
     & RealWorld & PAMAP2 & USC-HAD & HAD-AW & MyoGym   \\
    \toprule 
    \multicolumn{6}{c}{Random Forest Classifier}\\
    \midrule
    GPT-3.5 & \cellcolor[HTML]{caebc0}\textbf{79.70 $\pm$ 0.38} &  \cellcolor[HTML]{caebc0}\textbf{69.20 $\pm$ 0.29} & \cellcolor[HTML]{caebc0}\textbf{49.72 $\pm$ 0.67} &  48.98 $\pm$ 0.11  &  47.93 $\pm$ 0.58\\
    GPT-4 &  79.41 $\pm$ 0.29 &  67.73 $\pm$ 0.70 &  49.07 $\pm$ 0.27 &  48.61 $\pm$ 0.20&  47.64 $\pm$ 0.17 \\
    LLaMa 2  & 79.02 $\pm$ 1.01 &  67.82 $\pm$ 0.30&  49.41 $\pm$ 0.28 &  48.40 $\pm$ 0.08 &  47.95 $\pm$ 0.25 \\
    Palm 2 (Bard) &  78.79 $\pm$ 0.65  &  68.42 $\pm$ 0.43 & 49.06 $\pm$ 0.31 &  49.36 $\pm$ 0.05 &  47.91 $\pm$ 0.25 \\
    Gemini &  78.93 $\pm$ 0.17 &  67.53 $\pm$ 0.10 &  49.65 $\pm$ 0.51 &  48.70 $\pm$ 0.06 &  \cellcolor[HTML]{caebc0}\textbf{47.98 $\pm$ 0.22} \\
    \hline
    Real Data &  71.53 $\pm$ 1.07 &  67.08 $\pm$ 0.53&  47.23 $\pm$ 0.08 & \cellcolor[HTML]{caebc0}\textbf{52.77 $\pm$ 0.07} & 46.61 $\pm$ 0.09 \\
    
    \midrule
    \multicolumn{6}{c}{Deep ConvLSTM }\\
    \midrule
    GPT-3.5 & $82.25 \pm 0.32$ &  \cellcolor[HTML]{caebc0}\textbf{75.16 $\pm$ 0.82} &  61.44 $\pm$ 0.49 &  51.98 $\pm$ 0.28 &  49.46 $\pm$ 0.88\\
    GPT-4 &  80.20 $\pm$ 0.88&  73.59 $\pm$ 0.75 &  61.43 $\pm$ 0.59 &  51.11 $\pm$ 0.28 &  46.85 $\pm$ 0.28\\
    LLaMa 2  &  \cellcolor[HTML]{caebc0}\textbf{82.33 $\pm$ 0.34} &  74.12 $\pm$ 0.88 &  60.93 $\pm$ 0.39 &  50.94 $\pm$ 0.17 &  48.78 $\pm$ 0.45 \\
    Palm 2 (Bard) &  81.74 $\pm$ 0.48 &  74.66 $\pm$ 0.96 & 61.17 $\pm$ 0.44  &  51.03 $\pm$ 0.11 &  49.00 $\pm$ 0.39 \\
    Gemini &  80.86 $\pm$ 0.67 &  72.76 $\pm$ 0.37 &  61.02 $\pm$ 0.15 &  51.44 $\pm$ 0.49&  48.82 $\pm$ 0.31\\
    \hline
    Real Data &  77.79 $\pm$ 0.85 &  69.26 $\pm$ 1.07 & \cellcolor[HTML]{caebc0}\textbf{63.35 $\pm$ 0.67} & \cellcolor[HTML]{caebc0}\textbf{56.16 $\pm$ 0.38} &  \cellcolor[HTML]{caebc0}\textbf{50.69 $\pm$ 0.61}\\
    
    \midrule
    \multicolumn{6}{c}{Deep ConvLSTM with self attention}\\
    \midrule
    GPT-3.5 &  80.47 $\pm$ 0.49 &  73.45 $\pm$ 0.85 &  59.12 $\pm$ 0.62 &  47.67 $\pm$ 0.73 &  47.35 $\pm$ 0.41\\
    GPT-4 &  80.31 $\pm$ 0.36 &  73.68 $\pm$ 1.26 &  57.97 $\pm$ 0.68&  47.64 $\pm$ 0.50&  46.03 $\pm$ 0.29 \\
    LLaMa 2  &  80.36 $\pm$ 0.67 & \cellcolor[HTML]{caebc0}\textbf{73.77 $\pm$ 0.85} &  59.09 $\pm$ 0.89&  48.75 $\pm$ 1.07&  47.58 $\pm$ 0.47\\
    Palm 2 (Bard) &  80.77 $\pm$ 0.72 &  73.23 $\pm$ 0.92 & 58.66 $\pm$ 1.01 &  47.03 $\pm$ 0.90 &  47.21 $\pm$ 0.56 \\
    Gemini & \cellcolor[HTML]{caebc0}\textbf{80.82 $\pm$ 0.61} &  72.89 $\pm$ 0.34 &  58.84 $\pm$ 0.33 &  47.99 $\pm$ 0.06 &  47.10 $\pm$ 0.28 \\
    \hline
    Real Data &  77.50 $\pm$ 0.76 &  64.36 $\pm$ 0.52 & \cellcolor[HTML]{caebc0}\textbf{61.82 $\pm$ 0.82} & \cellcolor[HTML]{caebc0}\textbf{56.51 $\pm$ 0.19} &  \cellcolor[HTML]{caebc0}\textbf{50.63 $\pm$ 0.88} \\
    
\end{tabular}
    \label{tab:llm_eval}
\end{table}

%% file: sections/experiment/motion_eval.tex
\subsection{Impact of Motion-Synthesis Models on HAR}
\label{sec: motion_eval}

In addition to experimenting with different LLMs for generating textual descriptions of activities, we also experimented with using a variety of motion synthesis models to convert the textual descriptions into human motion sequences in order to determine how different generated motion sequences affect downstream activity recognition.

\subsubsection{Experiment Setting}
We convert 1,000 textual descriptions, generated by GPT-3.5, to motion sequences by using the following four motion synthesis models: 
\textit{i)} T2M-GPT \cite{zhang2023generating};
\textit{ii)} MotionGPT \cite{jiang2023motiongpt};
\textit{iii)} MotionDiffuse \cite{zhang2022motiondiffuse}; and 
\textit{iv)} ReMoDiffuse \cite{zhang2023remodiffuse}. 
Both T2M-GPT and MotionGPT are Variational-Auto-Encoder (VAE)-based pipelines. In these VAE-based pipelines, the input textual descriptions are tokenized using a text tokenizer. 
The text tokens are then used by a transformer to autoregressively generate motion tokens, which are converted into motion sequences using a motion decoder. 
On the other hand, MotionDiffuse and ReMoDiffuse are diffusion-based models. 
In these models, the encoded textual descriptions are used in a reverse diffusion process, where Gaussian noise is gradually denoised to produce motion sequences.
 
The main difference between VAE-based and diffusion-based models lies in the length of the generated motion. 
In VAE-based models, a trained stopping token exists, and the generation process stops after this token is generated, allowing motion sequences to vary in length depending on the input textual description. 
However, diffusion-based models do not have a stopping token, so all generated motion sequences have a predefined fixed length. 
In our experiments, we set the motion length to be six seconds, following previous work \cite{zhang2023remodiffuse}, for the diffusion-based models. 
With a fixed length, we observed that some generated motion sequences ended prematurely, as six seconds was not sufficient to portray the motion defined in the input textual description. 
Conversely, for some motion sequences, six seconds proved too long, resulting in repeated or miscellaneous motions at the end of the sequence. 
We did not encounter such issues with VAE-based models, as the generated motion sequences ended naturally.

The sizes of the real IMU dataset and the virtual IMU datsets are shown in \autoref{tab:motion_data_size} of \autoref{app:dataset_size}.  The classification models are trained on either real IMU data alone or on both the real IMU data and the virtual IMU data generated by IMUGPT using the four motion synthesis models.
 

\subsubsection{Results}
\ 
\input{tables/motion_eval}

\autoref{tab:motion_eval} shows the downstream activity recognition results when different motion synthesis models are used in IMUGPT for virtual IMU data generation. 
Overall, we observe that T2M-GPT generally performs better for most of the datasets and across different classifier models. 
With the Random Forest classifier, the downstream performance of T2M-GPT shows an
overall improvement of 0.7\% over MotionGPT, 1.2\% over MotionDiffuse, and 2.2\% over ReMoDiffuse across all the datasets (all differences are statistically significant). 
In some instances, other motion synthesis models return better results mainly with the DeepConvLSTM model with self-attention. 
However, for MyoGym, ReMoDiffuse appears to be the better motion synthesis model across all the three classifier models. 
Due to the complex activities in HAD-AW, its results are worse than for real data. 

%% file: tables/motion_eval.tex
\begin{table}[t]
  \centering
  \caption{Model performance (Macro F1) when trained on both the real IMU data and the virtual IMU data when various motion synthesis models are used to generate the motion sequences in IMUGPT. 
  ``Real Data'' denotes the baseline experiments not including any generated, virtual IMU data. 
  }
    \small
  \begin{tabular}{l c c c c c}
     & RealWorld & PAMAP2 & USC-HAD & HAD-AW & MyoGym   \\
    \toprule
    \multicolumn{6}{c}{Random Forest Classifier}\\
    \midrule
    T2M-GPT & \cellcolor[HTML]{caebc0}\textbf{79.70 $\pm$ 0.38} &  \cellcolor[HTML]{caebc0}\textbf{69.20 $\pm$ 0.29} & \cellcolor[HTML]{caebc0}\textbf{49.72 $\pm$ 0.67} &  48.98 $\pm$ 0.11  &  47.93 $\pm$ 0.58\\
    MotionGPT &  79.19 $\pm$ 0.71 & 68.00 $\pm$ 0.28 & 49.72 $\pm$ 0.37 & 48.74 $\pm$ 0.03 &  47.73 $\pm$ 0.62\\
    MotionDiffuse & 79.23 $\pm$ 0.40 &  68.15 $\pm$ 0.43 &  48.91 $\pm$ 0.49&  49.19 $\pm$ 0.09&  46.77 $\pm$ 0.21 \\
    ReMoDiffuse & 74.32 $\pm$ 0.87 & 67.81 $\pm$ 0.11 & 46.72 $\pm$ 0.30 &  50.05 $\pm$ 0.06 &  \cellcolor[HTML]{caebc0}\textbf{49.31 $\pm$ 0.27} \\
    \hline
        Real Data &  71.53 $\pm$ 1.07 &  67.08 $\pm$ 0.53&  47.23 $\pm$ 0.08&  \cellcolor[HTML]{caebc0}\textbf{52.77 $\pm$ 0.07} &  46.61 $\pm$ 0.09 \\
    \midrule
    \multicolumn{6}{c}{Deep ConvLSTM }\\
    \midrule
    T2M-GPT& \cellcolor[HTML]{caebc0}\textbf{82.25 $\pm$ 0.32} &  \cellcolor[HTML]{caebc0}\textbf{75.16 $\pm$ 0.82} &  61.44 $\pm$ 0.49 &  51.98 $\pm$ 0.28 &  49.46 $\pm$ 0.88\\
    MotionGPT & 81.45 $\pm$ 0.75 & 73.39 $\pm$ 0.53 &  60.72 $\pm$ 0.73&  50.49 $\pm$ 0.15 & 46.35 $\pm$ 0.48 \\
    MotionDiffuse &  82.03 $\pm$ 0.68 &  74.44 $\pm$ 0.75 & 61.07 $\pm$ 0.30 &  53.00 $\pm$ 0.30 &  46.78 $\pm$ 0.23 \\
    ReMoDiffuse &  78.61 $\pm$ 0.45 & 73.80 $\pm$ 0.58 &  61.11 $\pm$ 0.86 &  53.49 $\pm$ 0.36 &  \cellcolor[HTML]{caebc0}\textbf{51.93 $\pm$ 0.57} \\
    \hline
    Real Data &  77.79 $\pm$ 0.85 &  69.26 $\pm$ 1.07 &  \cellcolor[HTML]{caebc0}\textbf{63.35 $\pm$ 0.67} & \cellcolor[HTML]{caebc0}\textbf{56.16 $\pm$ 0.38} & 50.69 $\pm$ 0.61\\
    \midrule
    \multicolumn{6}{c}{Deep ConvLSTM with self attention }\\
    \midrule
    T2M-GPT &  80.47 $\pm$ 0.49 & \cellcolor[HTML]{caebc0}\textbf{73.45 $\pm$ 0.85} &  59.12 $\pm$ 0.62 &  47.67 $\pm$ 0.73 &  47.35 $\pm$ 0.41\\
    MotionGPT & 80.31 $\pm$ 0.47 & 72.20 $\pm$ 0.97 & 58.82 $\pm$ 0.59 & 46.96 $\pm$ 0.73 & 44.99 $\pm$ 0.51 \\
    MotionDiffuse & \cellcolor[HTML]{caebc0}\textbf{81.58 $\pm$ 0.22} &  73.29 $\pm$ 0.14 &  59.71 $\pm$ 0.58 &  51.19 $\pm$ 0.44 &  45.66 $\pm$ 0.09\\
    ReMoDiffuse &  77.65 $\pm$ 0.16 & 72.17 $\pm$ 1.24 &  60.71 $\pm$ 0.44 &  50.08 $\pm$ 0.82 &  48.49 $\pm$ 0.80 \\
    \hline
    Real Data &  77.50 $\pm$ 0.76 &  64.36 $\pm$ 0.52 & \cellcolor[HTML]{caebc0}\textbf{61.82 $\pm$ 0.82} & \cellcolor[HTML]{caebc0}\textbf{56.51 $\pm$ 0.19} & \cellcolor[HTML]{caebc0}\textbf{50.63 $\pm$ 0.88} \\
    
\end{tabular}
    \label{tab:motion_eval}
\end{table}

%% file: sections/experiment/diversity.tex
\subsection{Diversity as a Predictor for HAR Model Performance}
\label{sec: diversity_predictor}
We made two assumptions that form the basis for drawing a connection between the diversity of text and the downstream HAR performance. 
Our first hypothesis is that diversity in textual descriptions is correlated to diversity of the motion sequences generated by those descriptions. 
The second hypothesis is that diversity in motion sequences is correlated to performance for models trained on the virtual data obtained from those sequences. 
We validate these hypotheses by computing the correlation between the diversity of the textual descriptions and the motion sequences, and subsequently between motion sequences and final classifier performance. 
We also evaluate whether there exists a correlation between the textual descriptions and the model performance. 

\subsubsection{Correlations}
Using comparative diversity (section \ref{sec:comparative_diversity}), we compute the Pearson correlation coefficient between text diversity, motion diversity, and the change in the F1 score downstream. 
This process is similar to our saturation point identification algorithm (\autoref{alg:sat_point_ind}). 
For each activity, we start with a set of 50 text descriptions, adding 5\% more textual descriptions at each step, and calculate the MMD between the two sets. 
Corresponding motion sequences and virtual IMU data are generated at each step. 
For the motion sequences, we calculate the MMD in a similar manner. 
We then use the virtual IMU data to train classification models, as in previous experiments, and compare the performance to models trained on only real IMU data to obtain the per-class change in F1 score. 
This process is repeated until reaching the saturation point (\autoref{alg:sat_point_ind}). 
Ultimately, we have a list of values for text diversity, motion diversity, and changes in F1 scores, and we compute the correlation between these three factors.

For each dataset, we show the average correlation across all activities (\autoref{tab:correlation}). We find that text and motion diversity are highly correlated. For all but one dataset, HAD-AW, there is a negative moderate to strong correlation between the diversities and the downstream change in F1 score. We note that the correlation is negative because a lower MMD indicates higher diversity. This suggests that the diversity metric can serve as a downstream performance predictor, where higher diversity indicates better performance. Specifically, the RealWorld dataset showed the strongest correlation, as the virtual IMU data contributed to improvements in the per-class F1 score across all activities. In contrast, for the USC-HAD, PAMAP2, and MyoGym datasets, the virtual IMU data led to a decline in the per-class F1 score for some activities, resulting in a more moderate correlation. For the HAD-AW dataset, a positive correlation was observed, indicating that the virtual IMU data led to a drop in the per-class F1 score for more activities than it helped.

\input{tables/correlation}




\subsubsection{Saturation Point Identification Algorithm Evaluation}
\label{sec: algo_eval}

In order to evaluate our saturation point determination algorithm (\autoref{alg:sat_point_ind}), we use it to generate text descriptions for each activity across all the datasets. 
Our goal is to evaluate the set of descriptions generated by utilizing the algorithm, as opposed to the case when it is not used (i.e., textual descriptions are generated directly without accounting for diversity saturation).  
Therefore, we use the results from section \ref{sec: LLM_eval}, where 1,000 textual descriptions were generated, as a baseline. 
Consequently, we obtain two sets of text descriptions for each activity across all datasets: one set consisting of 1,000 descriptions and another set generated using \autoref{alg:sat_point_ind}, which includes descriptions produced up to the saturation point.

Subsequently, we generate virtual IMU data from these sets of textual descriptions, combine them with the corresponding real datasets, and use this data to train downstream classifiers. 
Downstream activity recognition results are shown in \autoref{tab:algo_eval}. 
We observe that across all activities,  datasets, and classifiers the performance in the Saturation Point case is similar to, if not better than, without using saturation point.
Note that the saturation points are computed for each activity individually. 
Most of the saturation points across all activities and datasets fall within the range of around 400 to 600 textual descriptions, indicating that the algorithm stops generation after that point. 
This suggests that we can achieve equivalent performance while utilizing around 50\% less data compared to directly generating 1,000 descriptions. 
Therefore, we can save at least 50\% of the time and compute resources needed for data generation. 
This algorithm also provides a formal structure to guide the process of generation in a manner that makes it consistent and repeatable -- in the absence of this algorithm, it would be impossible to determine how much textual descriptions should be generated, and at which point should generation be halted.  


\input{tables/algo_eval}

%% file: tables/correlation.tex
\begin{table}[t]
\centering
    \caption{ Correlations between comparative diversity of textual prompts, motion sequences, and the change downstream activity recognition perfomance (measured in macro F1 scores).}
    \begin{adjustbox}{width=\columnwidth,center}
    \begin{tabular}{c|c|c|c|c|c}
        Dataset & RealWorld & PAMAP2 & USC-HAD & HAD-AW & MyoGym \\
        \hline\hline
        Text V.S. Motion & $r = 0.92, p \leq 0.001$ & $r = 0.87, p \leq 0.001$ & $r = 0.91, p \leq 0.001$ & $r = 0.88, p \leq 0.001$ & $r = 0.91, p \leq 0.001$\\
        \hline
        Text V.S. F1 & $r = -0.77, p  \leq 0.001$ & $r = -0.31, p = 0.0052$ & $r = -0.45, p = 0.004$ & $r = 0.22, p = 0.173$ & $r = -0.24, p = 0.136$\\
        \hline
        Motion V.S. F1 & $r = -0.76, p \leq 0.001$ & $r = -0.32, p = 0.044$ & $r = -0.46, p = 0.003$ & $r = 0.21, p = 0.193$ & $r = -0.24, p = 0.136$\\
        \hline
    \end{tabular}
    \end{adjustbox}
    \label{tab:correlation}
\end{table}

%% file: tables/algo_eval.tex
\begin{table}[t]
  \centering
  \caption{Comparison of Model Performance (Macro F1 Score) Using Real and Virtual IMU Data, with and without the saturation Point Identification Algorithm. 
  }
  \begin{adjustbox}{width=0.9\textwidth}
  \begin{tabular}{l|c|c|c|c|c}
     & RealWorld & PAMAP2 & USC-HAD & HAD-AW & MyoGym   \\
    \toprule
    \multicolumn{6}{c}{Random Forest Classifier}\\
    \midrule
        Without Saturation Point  & $79.70 \pm 0.38$ &  69.20 $\pm$ 0.29 & 49.72 $\pm$ 0.67 &  48.98 $\pm$ 0.11  &  47.93 $\pm$ 0.58\\
    \hline
    With Saturation Point & 80.39 $\pm$ 0.34 &  69.50 $\pm$ 0.54 & 50.07 $\pm$ 0.10 &  50.62 $\pm$ 0.05 & 48.73 $\pm$ 0.15 \\
    \midrule
    \multicolumn{6}{c}{Deep ConvLSTM }\\
    \midrule
    Without Saturation Point & $82.25 \pm 0.32$ &  75.16 $\pm$ 0.82 &  61.44 $\pm$ 0.49 &  51.98 $\pm$ 0.28 &  49.46 $\pm$ 0.88\\
    \hline
    With Saturation Point & 82.70 $\pm$ 0.49 & 75.47 $\pm$ 1.15 & 61.36 $\pm$ 0.38 &  52.37 $\pm$ 0.27 & 48.74 $\pm$ 0.36 \\
    \midrule
    \multicolumn{6}{c}{Deep ConvLSTM with self attention }\\
    \midrule
    Without Saturation Point  &  80.47 $\pm$ 0.49 &  73.45 $\pm$ 0.85 &  59.12 $\pm$ 0.62 &  47.67 $\pm$ 0.73 &  47.35 $\pm$ 0.41\\
    \hline
    With Saturation Point & 81.11 $\pm$ 1.40 & 73.01 $\pm$ 0.45 & 59.79 $\pm$ 0.26 & 49.64 $\pm$ 0.32 & 47.53 $\pm$ 0.41 \\
    
\end{tabular}
    \end{adjustbox}
    \label{tab:algo_eval}
\end{table}

%% file: sections/experiment/motion_filter.tex
\subsection{Motion Filter Evaluation}
\label{sec: motion_filter_eval}

In this section, we evaluate the effectiveness of the newly introduced motion filter with regards to  eliminating irrelevant motion sequences, and how it affects the downstream activity recognition performance.

\subsubsection{Experimental Setting} We used the RealWorld dataset \cite{Sztyler2016realworld} for our evaluation. For each of the eight activities within the dataset, we first generated 50 textual descriptions of the activity using GPT-3.5 \cite{openai2022gpt}, and then used T2M-GPT \cite{zhang2023generating} to convert these descriptions into human motion sequences. We visualized these motion sequences as 3D animations. 
Some example visualizations are shown in \autoref{fig:sample_motion} of \autoref{app:sample_motion}. Using these visualizations, we manually annotated whether the sequences accurately portray the activity of interest. These manual annotations will serve as our ground truths for evaluating motion filtering performance.

The generated motion sequences are then processed with the motion filter to distinguish the relevance of the motion sequence. 
Specifically, we used GPT-3.5 and GPT-4 for our motion filter to further study the impact of LLMs on filtering performance.
Besides using LLMs for labeling motion captions, we manually annotated the captions generated by our motion caption model \cite{jiang2023motiongpt} to compare human and LLM performance 
The performance of our motion filter was evaluated with  
precision, recall, accuracy, F1 score, and percentage of incorrectly generated motion sequences before and after filtering.
For each LLM, we repeat the experiment five times on the same set of motion captions and report averages and standard deviations for each performance metrics. 







\input{tables/motion_filter_eval}

\subsubsection{Validation Results}
\autoref{tab:motion_filter_eval} shows the results for the motion filter using GPT-3.5, GPT-4, and human annotators to filter out incorrectly generated motion sequences using motion captions. The goal of the motion filter is to maximize true negatives (correctly identified inaccurate motion sequences) and minimize false positives (inaccurate sequences that escape the motion filter). Therefore, an effective motion filter should have high precision. We observe that the choice of LLM greatly impacts the motion filter's performance. GPT-4, with a precision of 0.901, significantly outperformed GPT-3.5, which had a precision of 0.706, comparable to human performance. When GPT-4 is used for filtering, the percentage of incorrectly generated motion sequences within the remaining dataset is greatly reduced, from 35.5\% before filtering to 9.9\% after. 

We note that both the LLMs and human annotators cannot achieve perfect performance in the filtering process using motion captions alone. This limitation stems from the fact that the motion captions generated by the motion caption model can sometimes be ambiguous and fail to accurately describe the input motion sequence. 

\subsubsection{Activity Recognition Performance}
\label{sec: motion_filter_perf}

We now examine the impact of the motion filter on the downstream activity recognition performance. 
We start with the motion sequences generated by T2M-GPT based on the text descriptions generated by GPT-3.5. 
The captions of the motion sequences are input into GPT-4, which filters out incorrectly generated motion sequences. 
\autoref{tab:filter_data_size} in \autoref{app:dataset_size} shows the size of the virtual IMU datasets after filtering.   
The classification models are then trained on the filtered datasets. 
The classification results, as detailed in \autoref{tab:motion_filter_f1}, indicate that using the motion filter leads to better downstream performance for the HAD-AW and MyoGym datasets, with relative performance improvements of 4.3\% and 4.1\% for HAD-AW and MyoGym, respectively, compared to not using the filter (all statistically significant differences).

However, for the other three datasets, the motion filters did not significantly impact downstream performance. 
This was surprising and contrary to our hypothesis and the motion filter validation results. 
We identified that motion filtering tends to reduce the diversity of the generated sequences, favoring similar kinds of motion sequences, potentially due to the biases in LLMs, which likely contributed to the observed decrease in performance.

\input{tables/motion_filter_f1}

%% file: tables/motion_filter_eval.tex
\begin{table}[t]
\centering
    \caption{Performance comparison of GPT-3.5, GPT-4, and human annotators in filtering out incorrectly generated motion sequences using motion captions. The percentage of incorrectly generated motion sequences before and after applying the motion filter is indicated as '\% before' and '\% after' the filter, respectively.
    }
\small
    \begin{tabular}{c|c|c|c|c|c|c}
    Annotator  & Precision & Recall & Accuracy & F1 & \% Before Filter & \% After Filter   \\
        \hline\hline

        GPT-3.5  & $70.56 \pm 0.84$& $71.49 \pm 2.01$& $59.95 \pm 1.01$ & $67.93 \pm 1.34$ & 35.50\% & $29.44\% \pm 0.84\%$\\
        \hline
        GPT-4   & $90.08 \pm 0.63$ & $60.55 \pm 2.08$ & $70.40 \pm 0.88$ & $71.12 \pm 1.46$ & 35.50\% & $9.92\% \pm 0.63\%$\\
        \hline
        Human   & 91.63 & 72.28 & 77.00 & 79.47 & 35.50\% & 8.37\%\\
        \hline
    \end{tabular}
    \label{tab:motion_filter_eval}
\end{table}



%% file: tables/motion_filter_f1.tex
\begin{table}[t]
  \centering
  \caption{Comparison of Model Performance (Macro F1) Using Real and Virtual IMU Data, with and without motion filter. 
  }
    \small
  \begin{tabular}{l|c|c|c|c|c}
     & RealWorld & PAMAP2 & USC-HAD & HAD-AW & MyoGym   \\
    \toprule
    \multicolumn{6}{c}{Random Forest Classifier}\\
    \midrule
    Without Motion Filter & $79.70 \pm 0.38$ &  69.20 $\pm$ 0.29 & 49.72 $\pm$ 0.67 &  48.98 $\pm$ 0.11  &  47.93 $\pm$ 0.58\\
    \hline
    With Motion Filter & 79.61 $\pm$ 0.44 &  70.22 $\pm$ 0.18 & 49.73 $\pm$ 0.14 &  51.09 $\pm$ 0.17 & 48.24 $\pm$ 0.19\\
    \midrule
    
    \multicolumn{6}{c}{Deep ConvLSTM }\\
    \hline
    Without Motion Filter & $82.25 \pm 0.32$ &  75.16 $\pm$ 0.82 &  61.44 $\pm$ 0.49 &  51.98 $\pm$ 0.28 &  49.46 $\pm$ 0.88\\
    \hline
    With Motion Filter & 81.45 $\pm$ 0.83 & 74.89 $\pm$ 1.14 & 61.75 $\pm$ 0.37 &  52.36 $\pm$ 0.20 & 51.94 $\pm$ 0.65 \\
    \midrule
    
    \multicolumn{6}{c}{Deep ConvLSTM with self attention }\\
    \hline
    Without Motion Filter &  80.47 $\pm$ 0.49 &  73.45 $\pm$ 0.85 &  59.12 $\pm$ 0.62 &  47.67 $\pm$ 0.73 &  47.35 $\pm$ 0.41\\
    \hline
    With Motion Filter & 81.66 $\pm$ 1.25 & 73.52 $\pm$ 0.46 & 60.67 $\pm$ 0.56 & 51.44 $\pm$ 1.10 & 50.50 $\pm$ 0.25 \\
    
\end{tabular}
    \label{tab:motion_filter_f1}
\end{table}

%% file: sections/discussion/main.tex
\section{Discussion}
Based on the investigations presented in this paper, we gain a range of insights into the overall process of language-based cross modality transfer and in IMUGPT 2.0 in particular.
We detail these in the subsequent sections, highlighting some notable aspects of the system and providing potential directions for future investigations. 

First, we study the effect of the presence of additional context information in the textual descriptions on the downstream classifier performance (section \ref{sec: prompt_enginnering}). Then, we investigate if the presence of multiple IMU sensors results in better performance gain from the addition of virtual IMU data (section \ref{sec:multiple_sensors}). Subsequently, we evaluate the effect of the motion filtering mechanism on the diversity of virtual IMU data (section \ref{sec:motion_filter_discussion}). Finally, we highlight the limitations of the extended system and offer some directions for future research (section \ref{sec:limit_future}).

\subsection{How much impact does context information have on downstream HAR performance?}
\label{sec: prompt_enginnering}


As reported in the results (\autoref{tab:llm_eval})
, while GPT-3.5 had the best overall performance for the datasets, the other LLMs were not far behind. This raises the question: \textit{To what extent does the context information within the generated textual descriptions affect downstream activity recognition performance?}  To evaluate this, we modified the prompts for GPT-3.5 to generate text descriptions with additional context information. The intuition behind using context information is that these factors could affect the actual motions in the activity performed, thereby generating varying virtual IMU data. 
Arguably, a wider range of contextual factors can influence the generation of textual descriptions.
While an exhaustive evaluation is challenging to conduct (and beyond the scope of this paper), we explore the, in our opinion, most important contextual parameters that may have an impact on generated textual descriptions, and thus, the generated movement data:

\begin{description}
    \item [Age:] The age of the person performing a particular activity could influence the motion (e.g., a toddler vs adult).

    \item [Physique:] Similar to age, the physique of a person could also affect the effort required and, hence, the motion of the activity being performed (e.g., skinny vs muscular).

    \item [Weather:] Weather conditions on a particular day can affect activities performed outdoors. Especially in extreme conditions, it could make performing the activity more challenging (e.g., sunny vs thunderstorm).
\end{description}

\input{tables/context_parameters}

We passed different combinations of these parameters as prompts to generate text descriptions, following the same pipeline as in our original experiments. Example textual descriptions can be found in \autoref{tab:more_prompts_context} of \autoref{app:text_descriptions}.
The results of the experiments with varying context information are displayed in \autoref{tab:context_parameters}. We observe no significant difference in downstream performance compared to the experiments without context information. This could be because the motion synthesis model is trained on motion capture data that are recorded indoors, so the weather-related context may not be represented in the generated motion sequences. Similarly, the textual descriptions within the HumanML3D dataset \cite{Guo_2022_CVPR}, on which the motion synthesis models are trained, do not contain additional information related to age and physique. Consequently, the generated motion sequences may not accurately reflect the differences in motion that these context parameters would introduce.


\subsection{Do multi-sensor setups benefit more from the use of virtual IMU data?} \label{sec:multiple_sensors}
In our experiments (\autoref{tab:llm_eval} and \autoref{tab:motion_eval}), we observe that the performance of downstream classifiers trained on the RealWorld and PAMAP2 datasets improves significantly with the addition of virtual IMU data. This trend holds across all three classifier models (i.e., Random Forest Classifier, Deep ConvLSTM and Deep ConvLSTM with self attention). 
Conversely, adding virtual IMU data does not equally improve the HAR performance for HAD-AW, and the performance for USC-HAD and MyoGym only improves for the Random Forest Classifier. 
One of the key differences between the datasets is the number of IMU sensors for each dataset. 
While RealWorld has six and PAMAP2 has three sensors, the USC-HAD, HAD-AW and MyoGym datasets only have a single sensor. 

If multiple sensors are used to record an activity, certain relationships would exist between all the sensors, since all the data would be drawn from the same motion. 
We attribute the improved performance to this factor -- adding virtual IMU data aids the downstream classifier in learning these relationships, which leads to better results. 
The datasets which contain just a single sensor are unable to benefit from this aspect since no such relationship can exist with just one sensor.   

%
%

\subsection{How does the motion filter affect data diversity?}
\label{sec:motion_filter_discussion}
In section \ref{sec: motion_filter_eval}, we observed that for three datasets--RealWorld, PAMAP2 and USC-HAD--the application of the motion filter does not lead to a significant improvement in downstream performance (Table \ref{tab:motion_filter_f1}), despite the fact that the motion filter is able to remove irrelevant motion sequences as shown in our evaluation (\autoref{tab:motion_filter_eval}). 

We hypothesize that while the filtering process does improve the relevance of the generated virtual IMU data, it also leads to a reduction in diversity. Both of which affect the downstream performance. As highlighted in section \ref{sec: diversity_predictor}, a lower diversity would lead to poorer performance. Therefore, we investigate the effect of motion filtering on the diversity of the motion sequences by utilizing the absolute diversity metrics discussed in section \ref{sec:abs_div_metric}. 
For each activity, we compute the standard deviation and centroid diversity metrics for the unfiltered motion sequences as well as the motion sequences retained after filtering. 

The diversity scores are listed in Table \ref{tab:motion_filter_diversity}. We observe that across all datasets and metrics, the diversity of the motion sequences decreases after the application of the filter. Thus, while the filtering process is able to improve the relevance of the virtual IMU data, it also brings down the diversity. This reduction in diversity counteracts the benefit provided by the increased relevance of the data. 
As a result, the downstream performance does not change significantly after motion filtering. 


\input{tables/motion_emb_diversity}

\subsection{Limitations and Future Work}
\label{sec:limit_future}

\subsubsection{Lack of Expressivity of Motion Synthesis Model}

As shown in our evaluation (\autoref{tab:llm_eval}), the generated virtual IMU data did not lead to improvements in downstream performance for the HAD-AW dataset. We identify two reasons for this: 
\textit{i)} the motion synthesis models were unable to generate some of the activities within the HAD-AW dataset; 
\textit{ii)} the used motion-to-sensor method, IMUSim \cite{Yound2011imusim}, is unable to capture the subtle movement characteristics present in the real IMU data.

Since the motion synthesis models are trained on the HumanML3D dataset, they can only generate motion sequences for the activities that are present within this dataset. Many activities in the HAD-AW, such as 'washing hands' and 'drawing', are not included in the HumanML3D dataset. This led to all the motion sequences generated for these activities being irrelevant. While the motion filter was able to filter out most of these irrelevant motion sequences, some remained in the filtered dataset, leading to worse downstream performance.

Recently, a new motion capture dataset, Motion-X \cite{lin2023motionx}, has been introduced. Motion-X contains three times more motion capture data than HumanML3D and includes a more diverse range of activities, both indoor and outdoor. We expect a motion synthesis model trained on the Motion-X dataset would generate motion sequences for a broader range of activities, potentially alleviating the problem of generating irrelevant motion sequences.

\subsubsection{Lack of Capturing Subtle Motions}

Many daily activities in the HAD-AW dataset, such as 'writing on paper' and 'typing on a keyboard', involve subtle movements at the sensor location (here: the wrist). 
These subtle movements result in nuanced characteristics within the real IMU data that IMUSim cannot capture through simulation, leading to a distribution difference between the real and virtual IMU data. This discrepancy also contributed to the deep learning classifiers overfitting to the virtual IMU data for some datasets. 
The classifiers learned features specific only to the virtual IMU data, rendering them not generalizable to the real IMU testing dataset. 
This problem was also observed in some vision-based cross modality transfer approaches, such as IMUTube \cite{Leng2023imutube}.

To address this challenge, Santhalingam et al.\ \cite{Santhalingam2023asl} trained a model to convert 3D hand poses into accurate virtual IMU data for sign language recognition. However, this model is specifically designed for wrist sensor locations and is not applicable to sensors located on other body parts. In our future work, we plan to collect a multi-modal dataset containing both motion capture and sensor data. This will enable us to train a model capable of accurately converting motion sequences to virtual IMU data for all body locations. With more accurate virtual IMU data, the deep learning classifiers can be trained to generalize better to real IMU data.

%% file: tables/context_parameters.tex
\begin{table}[t]
  \centering
  \caption{Performance (Macro F1) of GPT-3.5 on RealWorld dataset with T2M-GPT motion synthesis model when parameters are added to generate text descriptions with context information}
  \begin{adjustbox}{width=1\textwidth}
  
    \begin{tabular}{l|c|c|c|c|c|c|c|c}

    Model & No parameters & Age & Weather & Physique & Age \& & Age \& & Weather \& & Age, Weather   \\
     &  &  &  &  & Weather & Physique & Physique & \& Physique   \\
    \hline\hline
    Random Forest & 79.70 $\pm$ 0.38 & 79.09 $\pm$ 0.45 & 79.58 $\pm$ 0.87 & 79.27 $\pm$ 0.86  & 79.12 $\pm$ 0.43 & 79.48 $\pm$ 0.52 & 78.84 $\pm$ 0.30 & 79.32 $\pm$ 0.77 \\
    \hline
    Deep ConvLSTM & 82.25 $\pm$ 0.32 & 81.33 $\pm$ 0.50 & 80.64 $\pm$ 0.78 & 82.19 $\pm$ 0.56 & 81.20 $\pm$ 0.22 & 80.61 $\pm$ 0.25 & 81.00 $\pm$ 0.36 & 81.67 $\pm$ 0.60\\
    \hline
    Deep ConvLSTM self attention & 80.47 $\pm$ 0.49 & 80.90 $\pm$ 0.42 & 79.72 $\pm$ 0.70 & 80.50 $\pm$ 0.44 & 79.89 $\pm$ 0.36 & 79.80 $\pm$ 0.73 & 80.50 $\pm$ 0.47 & 80.94 $\pm$ 1.00\\
    \hline
    
\end{tabular}
    \end{adjustbox}
    \label{tab:context_parameters}
\end{table}

%% file: tables/motion_emb_diversity.tex
\begin{table}[t]
  \centering
  \caption{Comparison of absolute diversity metrics for motion embedding sequences, with and without motion filter}
  \begin{adjustbox}{width=0.85\textwidth}
  \begin{tabular}{l|c|c|c|c|c}
    Dataset & RealWorld & PAMAP2 & USC-HAD & HAD-AW & MyoGym   \\
    \toprule
    \multicolumn{6}{c}{Standard Deviation Metric}\\
    \midrule
    Without Motion Filter & 0.0487 & 0.0393 & 0.0392 & 0.0308 & 0.0397\\
    \hline
    With Motion Filter & 0.0358 & 0.0305 & 0.0307 & 0.0259 & 0.0312\\
    \midrule
    
    \multicolumn{6}{c}{Centroid Metric}\\
    \midrule
    Without Motion Filter & 0.0037 & 0.0027 & 0.0026 & 0.0014 & 0.0023 \\
    \hline
    With Motion Filter & 0.0021 & 0.0017 & 0.0017 & 0.0011 & 0.0014 \\
    
\end{tabular}
    \end{adjustbox}
    \label{tab:motion_filter_diversity}
\end{table}

%% file: sections/conclusion/main.tex
\section{Conclusion}



This paper presents a significant enhancement to the language-based cross modality transfer paradigm previously introduced and piloted in IMUGPT. 
We extend the IMUGPT system by introducing the motion filtering module to ensure the relevance of virtual IMU data and establishing metrics to measure the diversity of the generated data, which in turn helps determine how much data to generate. These improvements have been demonstrated to reduce the time and computational demands of data generation by over 50\%. 

We further consolidate the system with a comprehensive evaluation across various HAR datasets, LLMs, and motion synthesis models, confirming that using GPT-3.5 and T2M-GPT for virtual IMU data generation leads to the best downstream performance. 
IMUGPT 2.0 highlights the potential to develop more complex and generalizable HAR systems, thus alleviating a major hurdle in the field: the lack of large labeled datasets.

%% file: appendix/main.tex
\appendix
\section{Using LLM for activity filtering}
\label{app:motion_filter_LLM}

In section \ref{sec:llm_filter}, we used LLMs to filter out incorrectly generated motion sequences with motion captions as input. We provide the exact prompts that we used along with some example motion captions in \autoref{fig:motion_filter_llm}.

\input{figure/motion_filter_LLM}

\section{Datasets}
\label{app:datasets}
Our experimental evaluations are based on five benchmark datasets that are widely used in the HAR community. \autoref{tab:dataset} provides an overview.

The RealWorld dataset contains recordings of 15 subjects performing 8 activities, with sensors placed on 7 different locations on the body. Each activity was recorded for 10 minutes each (except for jumping which was recorded for 1.7 minutes), under realistic conditions. 

A total of 9 subjects performed 12 activities in the PAMAP2 dataset. Over 10 hours of data were collected from the subjects using 3 IMUs, 
and a heart rate monitor, sampled at 100Hz. The three IMUs were placed at the chest, the subject's dominant arm, and ankle of the subject's dominant side, while the heart rate monitor was placed at the chest.

For the USC-HAD dataset, 14 subjects recorded 5 trials for 12 activities each on different days at various indoor and outdoor settings. Given the location of the trails and the most common locations where people carry mobile phones, the front right hip was chosen as the location for the sensors. 

The HAD-AW dataset was created with 16 subjects wearing the Apple Watch Series One on the right wrist and recording 31 activities, but we are using only 25 activities since the other 6 activities are related to workouts in the gym and will be covered in the MyoGym dataset. The Apple Watch 
collects data from the subjects who repeat each activity 10 times, amounting to approximately 160 samples of each activity.

The MyoGym dataset is dedicated to 30 activities performed in the gym. 10 subjects recorded data performing a set of 10 repetitions of each activity using a Myo armband worn on the forearm. 
The exercises were done using free weights and targeted different muscle groups. 
For our experiments, we have downsampled the frequency to 20 Hz to match that of the virtual IMU data for all datasets.

\input{tables/dataset}

\clearpage

\section{Real and Virtual Dataset Sizes}
\label{app:dataset_size}

In this section, we provide the sizes of the real IMU datasets and virtual IMU datasets that we used in our experimental evaluation (section \ref{sec:experiment}). \autoref{tab:llm_data_size} shows the dataset sizes when different LLMs are used for the textual descriptions generation within IMUGPT. \autoref{tab:motion_data_size} shows the dataset sizes when different motion synthesis models are used to generate the motion sequences using the textual descriptions generated by GPT-3.5 as input. \autoref{tab:filter_data_size} shows the sizes of the virtual IMU datasets with and without using the motion filter.

\input{tables/llm_data_size}
\input{tables/motion_data_size}
\input{tables/filter_data_size}

\newpage

\section{Sample Generated Text Descriptions}
\autoref{tab:more_prompts} and \autoref{tab:more_prompts_context} gives an overview of some examples of textual descriptions generated through various LLMs. We use the following prompt template to generate textual descriptions of activities:
\begin{description}
    \item [System Message:] You are a prompt generator designed to generate textual description inputs for activities. Do not provide anything other than prompt
    \item [User Message:] Prompts must be less than 15 words, contain description of the action with no description for environment. The prompt should only describe one person and an adjective describing how the person does the activity. Here are some example inputs: 'a person puts their hands together, leans forwards slightly, then swings the arms from right to left',  'a man is practicing the waltz with a partner', 'a man steps forward and does a handstand',   'a man rises from the ground, walks in a circle', 'a person jogs in place, slowly at first, then increases speed.',  'a man playing violin', 'a man playing guitar', 'a man doing skip rope', 'a man walks happily'."  50 prompts for prompts for "activity".
\end{description}

\label{app:text_descriptions}

\input{tables/text_descriptions}

\input{tables/text_descriptions_context}

\clearpage

\section{Sample Generated Motion Sequences}
\label{app:sample_motion}

\input{figure/sample_motion}

%% file: figure/motion_filter_LLM.tex
\begin{figure*}[h]
    \centering
    \includegraphics[width=0.7\linewidth]{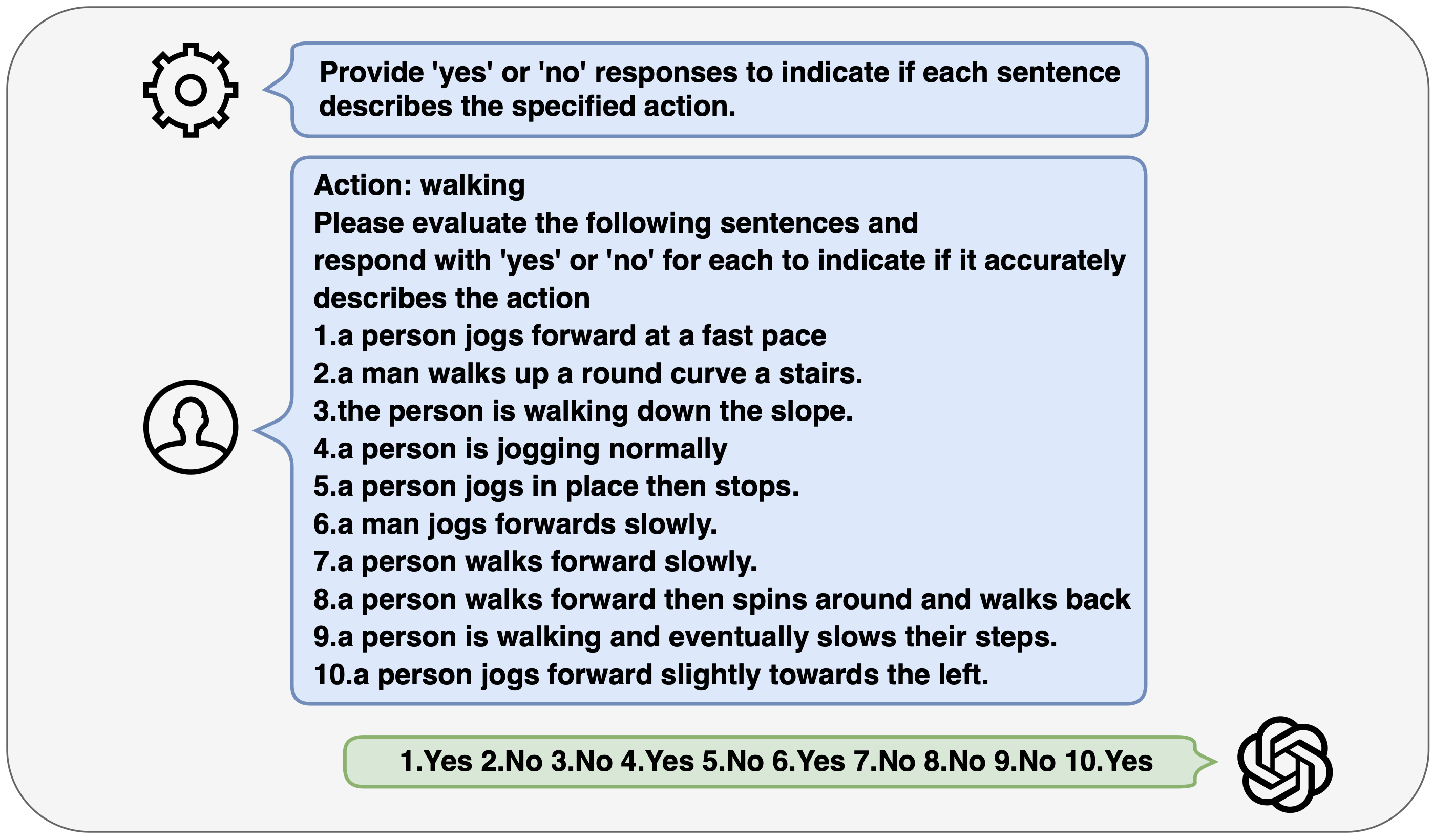}
    \vspace{-0.1in}
    \caption{ The prompts passed to the LLM for it to determine whether the given motion captions accurately describe the specified activity.
    }
    \label{fig:motion_filter_llm}
\end{figure*}

%% file: tables/dataset.tex
\begin{table}[h]
\centering
    \caption{ Summary of datasets. The percentage of samples for each activity is represented in the parentheses following the activities.
    }
    \begin{adjustbox}{width=\columnwidth,center}
    \begin{tabular}{c|p{1.5cm}|p{2.2cm}|p{1.4cm}|p{1.6cm}|p{8.5cm}}

        Dataset & Frequency & Sensor location & Subjects \# & Activities \# & Activities   \\
        \hline\hline

        RealWorld & 50 Hz & Head, Chest, Upper arm, Waist, Forearm, Thigh, Shin & 15 & 8 & climbing down (11.3\%), climbing up (13.2\%), jumping (2.1\%), lying (14.4\%), running (15.7\%), sitting (14.4\%), standing (14.2\%), walking (14.5\%) \\
        \hline

        PAMAP2 & 
        100 Hz & 
        Chest, Wrist, Ankle & 
        9 & 
        12 & 
        ascending stairs (6.1\%), cycling (8.5\%), descending stairs (5.4\%), ironing (12.3\%), lying (10.0\%), Nordic walking (9.7\%), rope jumping (2.2\%), running (5.1\%), sitting (9.6\%), standing (9.8\%), vacuum cleaning (9.1\%), walking (12.3\%) \\

        \hline

        USC-HAD & 100 Hz & Front Right Hip & 14 & 12 & elevator down (5.9\%), elevator up (8.5\%), jumping (3.8\%), running (6.3\%), sitting (9.3\%), sleeping (10.7\%), standing (8.4\%), walking forward (13.6\%), walking downstairs (7.0\%), walking left (9.2\%), walking right (9.8\%), walking upstairs (7.5\%) \\
        \hline

        HAD-AW & 50 Hz & Right wrist & 16 & 25 & Bed making (3.8\%), Cutting Components (4.2\%), Cycling (2.3\%), Dancing (3.0\%), Drawing (4.2\%), Driving Car (6.2\%), Eat Sandwich with Hand (3.5\%), Flipping (3.8\%), Playing on a violin (4.9\%), Playing on Guitar (3.6\%), Playing on Piano (5.3\%), Praying (5.2\%), Put off clothes (3.5\%), Reading (4.1\%), Rowing (3.8\%), Running (4.3\%), Shaking the dust (4.4\%), Showering (3.1\%), Sweeping (3.7\%), Typing on keyboard (4.1\%), Washing dishes (4.2\%), Washing hands (3.0\%), Wearing Clothes (3.6\%), Wiping (3.4\%), Writing on paper (4.6\%) \\
        \hline

        MyoGym & 50 Hz & Right forearm & 10 & 30 & 
        Bar Skullcrusher (3.7\%), Bench Dip / Dip (2.8\%), Bench Press (2.8\%), Bent Over Barbell Row (2.5\%), Cable Curl (3.1\%), Car Drivers (2.7\%), Close-Grip Barbell Bench Press (3.0\%), Concentration Curl (3.1\%), Dumbbell Alternate Bicep Curl (5.0\%), Dumbbell Flyes (4.3\%), Front Dumbbell Raise (4.7\%), Hammer Curl (4.3\%), Incline Dumbbell Flyes (4.2\%), Incline Dumbbell Press (3.7\%), Incline Hammer Curl (3.6\%), Leverage Chest Press (3.1\%), Lying Rear Delt Raise (2.9\%), One-Arm Dumbbell Row (3.0\%), Overhead Triceps Extension (3.1\%), Pushups (2.6\%), Reverse Grip Bent-Over Row (2.5\%), Seated Cable Rows (3.4\%), Seated Dumbbell Shoulder Press (3.1\%), Side Lateral Raise (3.1\%), Spider Curl (3.7\%), Tricep Dumbbell Kickback (2.8\%), Triceps Pushdown (3.0\%), Upright Barbell Row (3.0\%), Wide-Grip Front Pulldown (3.5\%), Wide-Grip Pulldown Behind The Neck (3.6\%) \\
    \end{tabular}
    \end{adjustbox}
    \label{tab:dataset}
\end{table}

%% file: tables/llm_data_size.tex
\begin{table}[h]
  \centering
  \caption{Real and virtual IMU data sizes when different LLMs are used for textual descriptions generation. ``Real Data'' denotes the baseline experiments not including any generated, virtual IMU data
  }
\small
  \begin{tabular}{l c c c c c}
     & RealWorld & PAMAP2 & USC-HAD & HAD-AW & MyoGym   \\
    \toprule 
    GPT-3.5 & 848 min & 1376 min & 1400 min & 3280 min  & 2231 min \\
    GPT-4 & 932 min & 1423 min & 1397 min & 3177 min  & 2862 min \\
    LLaMa 2 & 923 min & 1435 min & 1424 min & 3248 min  & 2549 min \\
    Palm 2 & 917 min & 1455 min & 1350 min & 3314 min  & 2392 min \\
    Gemini & 881 min & 1413 min & 1374 min & 3200 min  & 2367 min \\
    \hline
    Real Data & 1107 min & 322 min & 469 min & 662 min  & 154 min \\
    
\end{tabular}
    \label{tab:llm_data_size}
\end{table}

%% file: tables/motion_data_size.tex
\begin{table}[h]
  \centering
  \caption{Real and virtual IMU data sizes when different motion synthesis models are used. Motion sequences generated using textual descriptions generated by GPT-3.5. ``Real Data'' denotes the 
  }
\small
  \begin{tabular}{l c c c c c}
     & RealWorld & PAMAP2 & USC-HAD & HAD-AW & MyoGym   \\
    \toprule 
    T2M-GPT & 848 min & 1376 min & 1400 min & 3280 min  & 2231 min \\
    MotionGPT & 967 min & 1560 min & 1511 min & 3424 min  & 3234 min \\
    MotionDiffuse & 798 min & 1187 min & 1184 min & 2479 min  & 2975 min \\
    ReMoDiffuse & 798 min & 1195 min & 1195 min & 2479 min  & 2975 min \\
    \hline
    Real Data & 1107 min & 322 min & 469 min & 662 min  & 154 min \\
    
\end{tabular}
    \label{tab:motion_data_size}
\end{table}

%% file: tables/filter_data_size.tex
\begin{table}[h]
  \centering
  \caption{Comparison of virtual IMU dataset sizes with and without using motion filter
  }
    \small
  \begin{tabular}{l c c c c c}
     & RealWorld & PAMAP2 & USC-HAD & HAD-AW & MyoGym   \\
    \toprule 
    Without Motion Filter & 848 min & 1376 min & 1400 min & 3280 min  & 2231 min \\
    With Motion Filter & 330 min & 517 min & 417 min & 872 min  & 425 min \\
    
\end{tabular}
    \label{tab:filter_data_size}
\end{table}

%% file: tables/text_descriptions.tex
\begin{table}[h]
\centering
\vspace{0.25cm}
\caption{Example generated textual descriptions using various LLMs}
\begin{tabular}{|>{\scriptsize}p{0.055\columnwidth}|>{\scriptsize}p{0.085\columnwidth}|>{\scriptsize}p{0.85\columnwidth}|}
\hline
\multirow{5}{*}{walking} & GPT-3.5 & The person takes brisk steps, swinging their arms energetically \textbf{;} \quad Walking on a tightrope, the person carefully balances their body in mid-air \textbf{;} \quad A scientist paces back and forth in their lab, deep in thought, pondering a breakthrough \textbf{;} \quad Someone leisurely strolls down the tree-lined path, enjoying the serene surroundings \textbf{;} \quad A detective walks stealthily, investigating a crime scene for clues \\
\cline{2-3}
& GPT-4 & A woman strolls, pausing to admire the flowers  \textbf{;} \quad A jogger cools down, slowing to an easy walk \textbf{;} \quad A bride walks down the aisle, eyes locked on her partner \textbf{;} \quad She power-walks on the sunlit track field \textbf{;} \quad A janitor ambles around the school after hours \\
\cline{2-3}
& LLaMa 2 & A woman strolls casually, her hands in her pockets, enjoying the sunny day \textbf{;} \quad A woman power walks, her strides long and confident \textbf{;} \quad A man carries a heavy load, his steps slow and labored \textbf{;} \quad A child skips along the sidewalk, laughing with joy \textbf{;} \quad A person ambles through a museum, taking in the artwork \\
\cline{2-3}
& Palm 2 (Bard) & A person walks to the store \textbf{;} \quad A child skips down the sidewalk \textbf{;} \quad A person walks in a park, enjoying the grass \textbf{;} \quad A person walks alone \textbf{;} \quad A dog walks alongside its owner, wagging its tail \\
\cline{2-3}
& Gemini & A person walks slowly with a determined stride \textbf{;} \quad A soldier walks in formation, serving their country \textbf{;} \quad A relaxed man strolls in the park \textbf{;} \quad A child skip happily, a smile on their face \textbf{;} \quad A child skips down the sidewalk \\
\hline

\multirow{5}{*}{jumping} & GPT-3.5 & A person crouches down, then springs upward with both feet leaving the ground \textbf{;} \quad A child jumps on a trampoline, bouncing up and down with sheer delight \textbf{;} \quad Leaping over a stream, a hiker successfully clears the water, continuing their trek through the forest \textbf{;} \quad The person jumps with precision and accuracy, landing softly on the ground \textbf{;} \quad Performing a series of jumps, the person maintains a steady rhythm and flawless technique \\
\cline{2-3}
& GPT-4 & A woman crouches low, then springs up into a high jump \textbf{;} \quad A soccer player jumps up to head a high ball \textbf{;} \quad A frog enthusiast mimics a frog, leaping around a pond \textbf{;} \quad She jumps in place, warming up before her run \textbf{;} \quad Launching off one foot, he soars with ease \\
\cline{2-3}
& LLaMa 2 & A man jumps onto a fitness class, his legs spread wide apart \textbf{;} \quad A person jumps over a puddle, avoiding a splashy landing \textbf{;} \quad A woman jumps rope, her arms swinging in perfect rhythm \textbf{;} \quad A woman jumps over a hurdle, her legs long and lean \textbf{;} \quad A child jumps off a swing, their arms spread wide in excitement \\
\cline{2-3}
& Palm 2 (Bard) & A person jumps up and down in excitement \textbf{;} \quad A person jumps to be alive \textbf{;} \quad A person jumps to celebrate \textbf{;} \quad A person jumps to get exercise \textbf{;} \quad A person jumps to be confident \\
\cline{2-3}
& Gemini & A girl with arms up and legs straight leaps skywards \textbf{;} \quad A girl jumps up and down to a catchy song \textbf{;} \quad A boy jumps to catch a falling object \textbf{;} \quad A girl jumps to show off \textbf{;} \quad A cat jumps out of a window, thirsty for a taste of freedom \\
\hline

\multirow{5}{*}{cycling} & GPT-3.5 & A person mounts the bicycle, grips the handlebars, and pushes off with one foot \textbf{;} \quad A cyclist stands up on the pedals, exerting power to gain speed \textbf{;} \quad A cyclist maintains a steady pace, finding solace in the rhythmic motion \textbf{;} \quad Balancing on one wheel, the cyclist effortlessly rides a unicycle \textbf{;} \quad A cyclist accelerates, increasing their speed with each push of the pedal \\
\cline{2-3}
& GPT-4 & A cyclist weaves expertly through city traffic \textbf{;} \quad The biker uses hand signals when approaching a turn \textbf{;} \quad A racer crouches low, speeding on the track \textbf{;} \quad On a country road, the biker enjoys a solo ride at dawn \textbf{;} \quad He shifts gears, cycling up the steep incline \\
\cline{2-3}
& LLaMa 2 & A woman pedals up a steep hill, her face determined \textbf{;} \quad A person races against time, speeding through an empty park on their bike \textbf{;} \quad A man in a cycling competition takes a sharp corner, his tires screeching \textbf{;} \quad A person rides a bicycle through a scenic park, enjoying the views \textbf{;} \quad A person cycles through a bustling park, dodging pedestrians \\
\cline{2-3}
& Palm 2 (Bard) & A cyclist rides a bike down a winding road \textbf{;} \quad A cyclist rides a bike to improve their fitness \textbf{;} \quad A cyclist rides his bike in the heat \textbf{;} \quad A man rides a bicycle down a busy street \textbf{;} \quad A cyclist rides for freedom \\
\cline{2-3}
& Gemini & Man balances and propels forward on a bicycle with confidence \textbf{;} \quad A cyclist pedals vigorously up a steep hill \textbf{;} \quad An elderly man cycles slowly on a rural path \textbf{;} \quad A man in a suit cycles briskly to work \textbf{;} \quad A young boy cycles through the park, wind in his hair \\
\hline

\multirow{5}{*}{pushups} & GPT-3.5 & A person begins in a plank position, lowers their body with controlled movement \textbf{;} \quad A person lowers their body towards the ground, then pushes themselves back up \textbf{;} \quad A person's biceps contract as they lower their chest towards the floor \textbf{;} \quad A person maintains a steady pace, counting each pushup as they go \textbf{;} \quad On a mat, a woman extends and bends her arms, repeatedly \\
\cline{2-3}
& GPT-4 & Her triceps work hard as she raises up \textbf{;} \quad Arms quaking, he pushes through the burn \textbf{;} \quad He lowers his chest to just above the floor \textbf{;} \quad A hiker does pushups beside the trail \textbf{;} \quad A senior does modified pushups against a wall, building strength safely \\
\cline{2-3}
& LLaMa 2 & A person starts in a plank position, lowering their body until their chest nearly touches the ground \textbf{;} \quad A man uses a kettlebell to add weight to his pushups \textbf{;} \quad A man in a spin class does pushups, his legs still pedaling as he works his upper body \textbf{;} \quad A man in his 20s does pushups during a break at school or work \textbf{;} \quad A man grunts with effort as he completes a set of pushups \\
\cline{2-3}
& Palm 2 (Bard) & A person lowers their body until their chest touches the ground, then pushes back up to the starting position \textbf{;} \quad A person does a pushup with their feet on a chair \textbf{;} \quad A person does a pushup with their feet elevated \textbf{;} \quad A person does a pushup, then stands up \textbf{;} \quad A person does a push-up with their legs bent \\
\cline{2-3}
& Gemini & A young girl smoothly transitions from plank to pushup \textbf{;} \quad A woman with a fit physique does clapping pushups with ease \textbf{;} \quad A group of friends do pushups together as a challenge \textbf{;} \quad A woman does pushups on a bench \textbf{;} \quad A boy with short hair grunts as he does pushups \\
\hline
\end{tabular}

\label{tab:more_prompts}
\end{table}

%% file: tables/text_descriptions_context.tex
\begin{table}[h]
\centering
\vspace{0.25cm}
\caption{Example generated textual descriptions with context information using GPT-3.5 for 'walking'}
\begin{tabular}{|>{\scriptsize}p{0.1\columnwidth}|>{\scriptsize}p{0.9\columnwidth}|}
\hline
No parameters & The person takes brisk steps, swinging their arms energetically \textbf{;} \quad Walking on a tightrope, the person carefully balances their body in mid-air \textbf{;} \quad A scientist paces back and forth in their lab, deep in thought, pondering a breakthrough \textbf{;} \quad Someone leisurely strolls down the tree-lined path, enjoying the serene surroundings \textbf{;} \quad A detective walks stealthily, investigating a crime scene for clues \\
\hline
Age & A teenager walks with confidence, swinging their arms rhythmically  \textbf{;} \quad An elderly woman uses a walking cane to steady herself as she slowly moves forward \textbf{;} \quad A middle-aged woman walks briskly, multitasking by talking on her phone and checking her watch \textbf{;} \quad A baby takes independent steps, their face full of determination and triumph \textbf{;} \quad A school-aged child skips and hops, finding joy in every step they take \\
\hline
Weather & A person walks with an umbrella, shielding themselves from pouring rain \textbf{;} \quad A jogger perseveres through blistering heat, their body glistening with sweat \textbf{;} \quad A woman walks leisurely on a cool autumn evening as leaves softly carpet her path \textbf{;} \quad A man walks steadily through a gusty wind, his jacket billowing behind him \textbf{;} \quad In a snowdrift, a person walks hesitantly, lifting their legs with each step \\
\hline
Physique & A slender woman strides gracefully, her limbs moving in perfect synchronization \textbf{;} \quad A sturdy hiker navigates rough terrain, carefully placing each step to maintain balance \textbf{;} \quad A petite gymnast tumbles and flips while walking, turning the pathway into a personal gym mat \textbf{;} \quad A tall woman walks gracefully, elongating her strides with elegance \textbf{;} \quad A muscular man power-walks, propelling himself forward with each determined step \\
\hline
Age \& Weather & A young child takes small steps, holding their parent's hand while walking in the rain \textbf{;} \quad A woman in her forties walks briskly on a treadmill at the gym, headphones in \textbf{;} \quad A person walks carefully on an icy sidewalk, taking small steps to avoid slipping \textbf{;} \quad n elderly woman walks slowly with a cane, determined to stay active despite the rain \textbf{;} \quad A person walks on a sandy beach, feeling the warmth of the sun on their skin \\
\hline
Age \& Physique & A fit and athletic teenager strides confidently, covering a long distance in each step \textbf{;} \quad A young athlete jogs lightly in between walking, maintaining an energetic and athletic stride \textbf{;} \quad An elderly woman holds onto a walking stick, taking slow and measured steps \textbf{;} \quad An overweight man waddles slowly, his gait slightly unstable, breathing heavily \textbf{;} \quad A person with a prosthetic limb walks determinedly, their movements adapted but resilient \\
\hline
Weather \& Physique & A slender woman effortlessly glides through the rain, walking with graceful determination \textbf{;} \quad A person determinedly walks through a muddy field, pushing through the resistance \textbf{;} \quad A hunched elderly man shuffles slowly with the aid of a cane during a snowstorm \textbf{;} \quad A lean man power walks through a snowstorm, his breath visible in the freezing air \textbf{;} \quad A petite man walks briskly, his coat flapping in the strong breeze \\
\hline
Age, Weather \& Physique & An elderly woman strolls slowly, using a cane, under a clear blue sky \textbf{;} \quad A petite girl skips merrily, twirling her colorful umbrella on a rainy day \textbf{;} \quad A slim and athletic woman hikes uphill, sweat glistening on her forehead, on a sunny day \textbf{;} \quad A pregnant woman takes gentle walks around the neighborhood, supporting her growing belly \textbf{;} \quad A middle-aged man strolls leisurely along the beach, enjoying the warm breeze \\
\hline

\end{tabular}
\label{tab:more_prompts_context}
\end{table}

%% file: figure/sample_motion.tex
\begin{figure*}[h]
    \centering
    \includegraphics[width=0.73\linewidth]{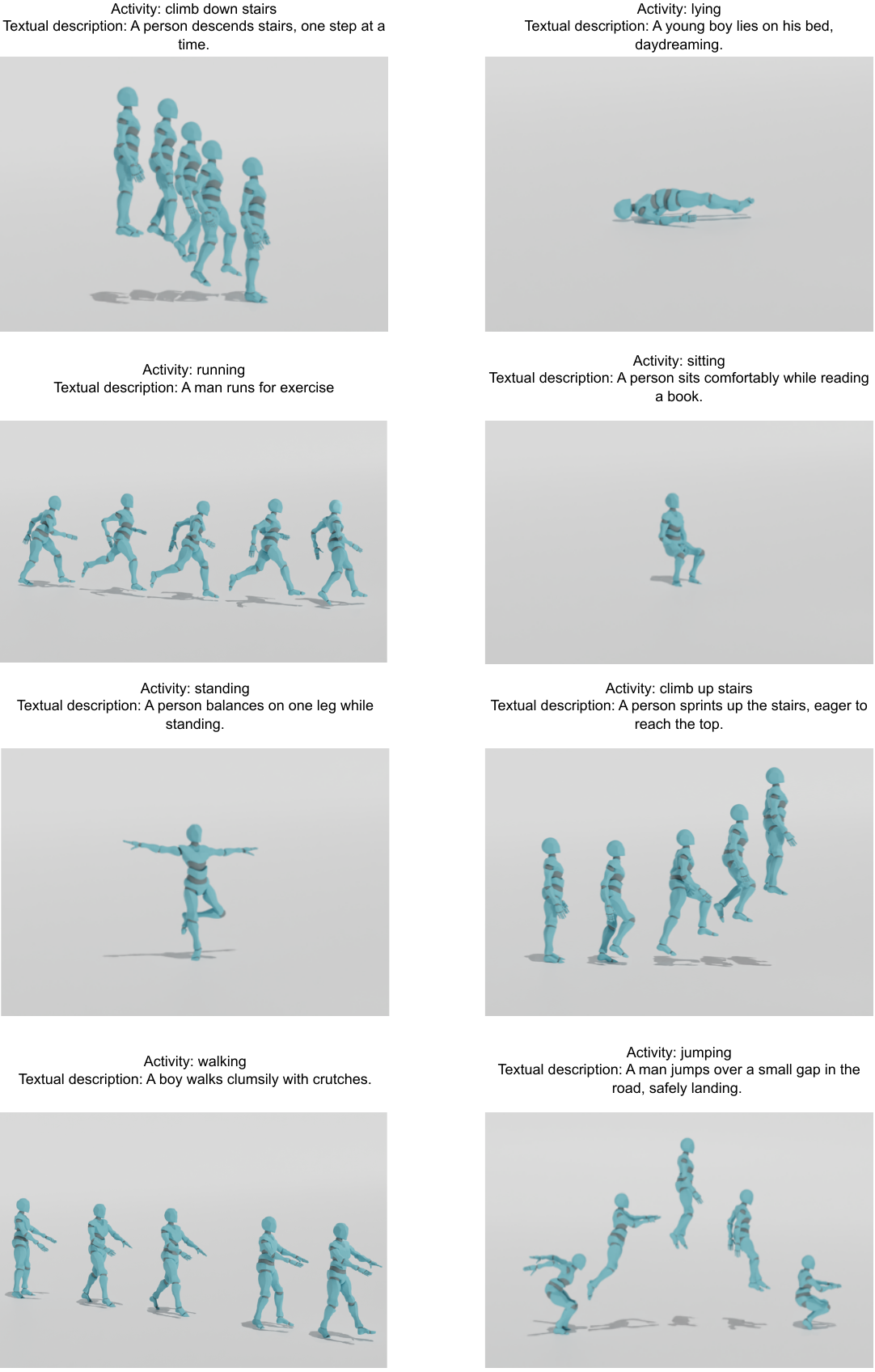}
    \vspace{-0.1in}
    \caption{Example visualized motion sequences for activities in the RealWorld dataset.
    }
    
    \label{fig:sample_motion}

\end{figure*}